\documentclass{article}

    \usepackage[final]{neurips_2020}

\usepackage[utf8]{inputenc} 
\usepackage[T1]{fontenc}    
\usepackage{hyperref}       
\usepackage{url}            
\usepackage{booktabs}       
\usepackage{amsfonts}       
\usepackage{nicefrac}       
\usepackage{microtype}      

\usepackage{wrapfig}
\usepackage{amsmath}
\usepackage{graphicx}
\usepackage{xcolor}
\usepackage{textcomp}

\usepackage{bm}
\usepackage{amssymb}
\usepackage{caption}
\usepackage{subcaption}

\DeclareMathOperator*{\argmax}{arg\,max}

\usepackage[longend]{algorithm2e}
\usepackage{textgreek}
\usepackage{tcolorbox}

\usepackage{hyperref}
\hypersetup{
    colorlinks=true,
    linkcolor=blue,
    filecolor=magenta,      
    urlcolor=cyan,
}
\title{Experimental design for MRI by greedy policy search}

\author{
  Tim Bakker,\\
  University of Amsterdam,\\
  \texttt{t.b.bakker@uva.nl} \\
  \And
  Herke van Hoof, \\
  University of Amsterdam \\ 
  \texttt{h.c.vanhoof@uva.nl} \\
  \And
  Max Welling, \\
  University of Amsterdam, CIFAR \\   
  \texttt{m.welling@uva.nl} \\
}

\begin{document}

\maketitle

\begin{abstract}
    In today's clinical practice, magnetic resonance imaging (MRI) is routinely accelerated through subsampling of the associated Fourier domain. Currently, the construction of these subsampling strategies - known as experimental design - relies primarily on heuristics. We propose to learn experimental design strategies for accelerated MRI with policy gradient methods. Unexpectedly, our experiments show that a simple greedy approximation of the objective leads to solutions nearly on-par with the more general non-greedy approach. We offer a partial explanation for this phenomenon rooted in greater variance in the non-greedy objective's gradient estimates, and experimentally verify that this variance hampers non-greedy models in adapting their policies to individual MR images. We empirically show that this adaptivity is key to improving subsampling designs.
\end{abstract}

\section{Introduction}
Magnetic resonance imaging (MRI) is a non-invasive medical imaging technique with a wide range of diagnostic applications. However, long acquisition times during the imaging process limit patient comfort, throughput, and imaging quality (for instance due to patient movement) \citep{fastmri}. Reducing imaging times has been an active field of research for the past fifty years. A potential avenue for tackling this problem is to acquire less measurement data during a scan, linearly reducing acquisition times: this is often referred to as accelerated MRI \citep{fastmri}. 

Measurements in MR imaging are performed in the frequency domain, also known as \textit{k-space}. These measurements are transformed (reconstructed) into the familiar MR images through the inverse Fourier transform. Accelerating MRI - reducing data acquisition - amounts to subsampling the k-space, which due to the Nyquist-Shannon sampling theorem will introduce aliasing artefacts in naive reconstructions. The presence of such artefacts renders the resulting images unusable for diagnostic purposes \citep{fastmri}: in order to improve image quality, additional information must be included in the reconstruction process. In clinical settings today, this is typically done using compressed sensing (CS) techniques \citep{cs, cs2, cs3}. With the rise of deep learning (DL), some successes have also been seen using deep reconstruction networks to obtain diagnostic quality images from more aggressively subsampled k-space \citep{jin17, rim2, hyun, schlemper, hammernik, wang, irim, irimmri, ete}. The additional information utilised is implicitly learned from training data. Such neural networks are trained by applying predetermined subsampling masks to k-space: from this masked frequency domain the model then learns to reconstruct target images obtained from the fully sampled k-space \citep{fastmri}. In these CS and DL settings, subsampling masks are typically determined beforehand: either carefully crafted by experts, or based on heuristics \citep{jin19}. 

A natural next step is a move away from handcrafted subsampling masks towards learned acquisition strategies. The process of choosing an optimal set of measurements is known as experimental design \citep{seegercs}. While such design methods can be employed to learn a fixed subsampling mask for a data set \citep{irismri, bahadir, weiss, gozcu, sanchez1, sanchez2}, an ostensively more salient approach is to learn an adaptive strategy that has the ability to propose different masks for different MR images (e.g. various patients or locations). Intuitively, such methods should outperform their non-adaptive counterparts on reconstruction quality given the same measurement budget. In the DL literature two trends are noticeable. The first learns to sequentially acquire measurements in k-space until some budget is exhausted. These adaptive acquisition methods are trained separately \citep{pineda} or jointly \citep{jin19} with the reconstruction model, based on some heuristic \citep{sanchez3}, or a combination of these \citep{zhang}. The second type of approach involves jointly learning an optimal subsampling and reconstruction by parameterising the mask itself \citep{weiss, bahadir, irismri}. These directly learned masks are typically not adaptive.

Here, we frame finding an optimal subsampling mask as a reinforcement learning (RL) problem. Our approach - like \citep{jin19} and (concurrently) \citep{pineda} - employs a separate reconstruction and policy model. The reconstruction model performs reconstructions from k-space measurements suggested by the policy model. The policy model suggests which measurement to make, based on the current reconstruction. The reward of each of these measure-reconstruct cycles is given as the improvement in reconstruction quality provided by some appropriate metric. 
\begin{figure} 
  \centering
  \includegraphics[width=0.8\linewidth]{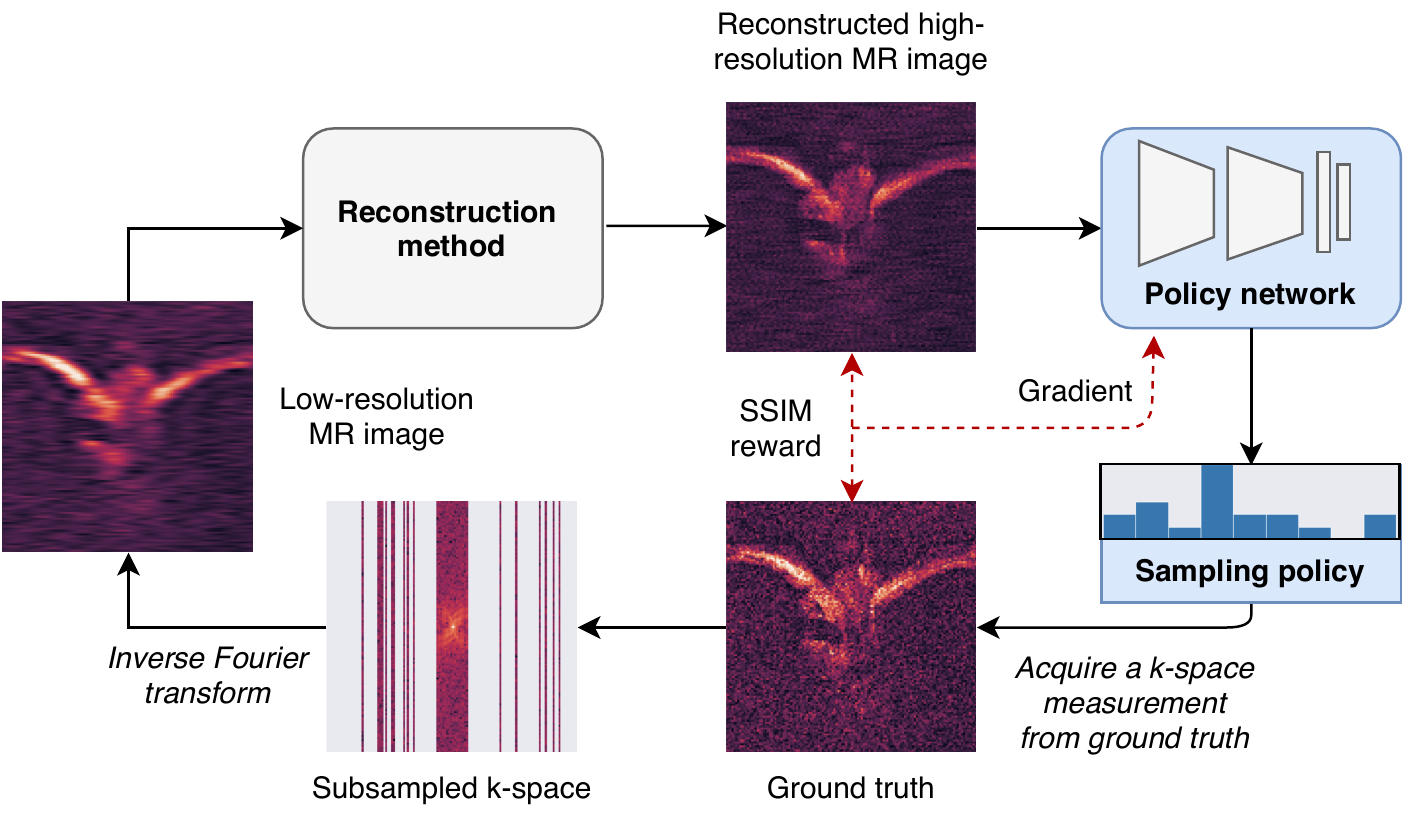}
  \caption{The iterative acquisition procedure. An initial subsampling of k-space is obtained from the ground truth image. The subsampled frequency domain is fed into a reconstruction method, which for neural network-based reconstructions typically starts with an inverse Fourier transform, such that the input and output domain match. This intermediate step results in a low-resolution, so-called \textit{zero-filled} reconstruction \cite{jin17}. The high-resolution reconstructed MR image is input to the policy network, which outputs a discrete probability distribution that represents the suggested sampling policy. An action is sampled from this policy, corresponding to a measurement of k-space. This measurement is simulated from the ground truth MR image, and the procedure is repeated until the acquisition budget is exhausted. The reward of an acquisition step is given by the improvement in SSIM of the ground truth and reconstruction resulting from that acquisition.}
  \label{fig:loop}
\end{figure}
The reinforcement learning perspective additionally allows for an analysis of a second split in the literature: greedy and non-greedy models. When optimising an MRI subsampling mask - given some budget - a natural question to ask is whether direct optimisation of the long-term reconstruction quality enjoys strong advantages over a greedy optimisation that acquires the measurement that leads to the greatest immediate improvement.  

In the literature, proposed models are often compared to compressed sensing baselines, but rarely to other DL methods due to an historical lack of standardised datasets and evaluation metrics. This limits MRI experimental design research, as it is unclear how different families of methods (e.g. adaptive/non-adaptive, greedy/non-greedy) compare, and thus which directions are the most promising future research targets. We attempt to bridge some of this gap here, in the hope to guide future research into experimental design for MRI. In this paper, we provide a proof of concept showing that direct policy gradient methods can be used to learn experimental design for MRI. We construct both greedy and a non-greedy models using the above-mentioned RL framework. The greedy model is quicker to train and outperforms some of its non-greedy counterparts in our experiments. We hypothesise that the underperformance of non-greedy models is partially due to weaker adaptivity to individual MR slices as a consequence of greater variance in the gradient estimates, and we provide experimental evidence for this hypothesis.\footnote{Code is available at: \url{https://github.com/Timsey/pg_mri}.}

\section{Related work} \label{sec:rel}
With the introduction of compressed sensing, several methods for designing MRI undersampling masks emerged. The original theoretical justification for CS prescribes the use of uniformly random undersampling masks, incidentally removing the need for specific designs \citep{cs2, cs3}. However, most practical success was achieved using variable-density sampling (VDS) heuristics \citep{cs}, which employ higher sampling density in the low-frequency regions of the Fourier domain. We refer to \citep{csrev} for an overview. There has also been interest in data-driven mask design for compressed sensing \citep{oedipus, sherry, vella}. Such methods often design fixed (non-adaptive) masks that perform well on average over a particular data set. Some recent models in this area include \citep{gozcu, sanchez1, sanchez2}, which aim to select an optimal stochastic greedy mask for a fixed budget, using training data as guidance. These models can be applied on top of any reconstruction method, exploiting the recent surge of interest in developing strong (even pre-trained) deep reconstruction models for arbitrary mask designs \citep{jin17, rim2, hyun, schlemper, hammernik, wang, irim, irimmri, ete}. The work by \citep{seegeral, seegeral2} instead proposed to use the criterion of maximising posterior information gain to guide a greedy, adaptive mask design. 

While deep learning research for undersampled MRI reconstruction has primarily focused on improving reconstruction methods, there has been some success learning undersampling masks in recent years as well. The existing methods can roughly be characterised by three families: non-greedy and non-adaptive, greedy and adaptive, or non-greedy and adaptive. 

Notable non-greedy, non-adaptive strategies have been proposed by \citep{irismri, weiss, bahadir}. Introducing various continuous relaxations of the binary subsampling mask, these works jointly optimise their mask design with a reconstruction network. A greedy, adaptive method by \citep{sanchez3} estimates a posterior over undersampled MR reconstructions with a conditional Wasserstein GAN \citep{wgan}, and uses empirical pixel-wise variance of the Fourier domain under this posterior to guide an acquisition strategy. In a similar vein, \citep{zhang} learns a reconstruction network as well as a pixel-wise uncertainty measure according to the model proposed in \citep{kengal}. An evaluator network is adversarially trained alongside to learn to recognise mistakes in k-space (implicit in the reconstruction) made by the reconstruction network. The evaluator is used to guide the acquisition process, while the uncertainty measure is used for monitoring only. Joint training of the evaluator network with the reconstruction network is crucial, as the reconstruction network must be incentivised to produce reconstructions that have consistent k-space representation for evaluator based acquisition to perform well.\footnote{While we would ideally compare our method to this work, this is infeasible due to reasons discussed in~\ref{sec:zhang}.}

Unique in the literature is the non-greedy, adaptive method by \citep{jin19}. Inspired by DeepMind's AlphaZero \citep{alphazero}, the authors propose to jointly train a reconstruction and policy network, using Monte Carlo tree search to generate self-supervised targets for the sampling policy. Similar to our proposed method, the reconstruction and sampling models can be decoupled, and as such a sampling strategy can be learned for any given reconstruction method. Unlike our approach however, MCTS based training does not naturally allow for training greedy models. Additionally, our approach enjoys a computational advantage - due to the use of smaller models - and converges more quickly. Finally, direct policy optimisation involves fewer design choices than computing an MCTS distribution.


\section{Background}
\subsection{Policy gradients and reinforcement learning}
We formalise the sequential selection of subsampling masks for MRI as a Partially Observable Markov Decision Process (POMDP). The latent state $z_t$ at acquisition step $t$ of this POMDP corresponds to a tuple $(\bm{x}, h_t)$ of the true underlying MR image $\bm{x}$ and the history $h_t$ of actions $a$ and observations $o$: $h_t = (a_0, o_0, ..., a_{t-1}, o_{t-1})$. An action $a$ represents a particular k-space measurement, and the corresponding observation $o$ is the result of that measurement. The agent takes as internal state a summary of the history $h_t$, provided by the current reconstruction $\hat{\bm{x}}_t$, and outputs a policy $\pi(a_t|\hat{\bm{x}}_t)$ over actions. The reward $r(z_t, a_t)$ is the improvement in reconstruction quality due to measurement $a_t$ taken when the latent state is $z_t$. The goal here is to learn a policy that maximises the expected sum of rewards (i.e. return) given some measurement budget.



Policy gradient methods are an approach for directly maximising the expected return $J(\phi)$ of a policy $\pi_\phi$ parameterised by $\phi$ on such a POMDP. The log-ratio trick can be used to rewrite the gradient of $J$ with respect to the policy parameters $\phi$ as an expectation of a gradient \citep{sutton, gpomdp}:
\begin{equation} \label{eq:pg}
    \nabla_\phi J(\phi) = \mathbb{E}_{\hat{\bm{x}}_0} \sum_{t=0}^{T-1} \left[ \nabla_\phi \log \pi_\phi(a_t|\hat{\bm{x}}_t) \sum_{t'=t}^{T-1} \gamma^{t'-t} \left( r(z_{t'}, a_{t'}) - b(z_{t'}) \right) \right].
\end{equation}
Here, $\mathbb{E}_{\hat{\bm{x}}_0}$ is an expectation over initial states $\hat{\bm{x}}_0$, and $t \in [0, T-1]$ indexes the time step of an episode. The introduction of a reward baseline $b(z_{t'})$ reduces the typically high variance of this estimator \citep{sutton, williams}. This baseline can be any function that is independent of the choice of action $a_{t'}$. In our setting, we will construct $b(z_{t'})$ out of rewards obtained from multiple rollouts that start in the same state $z_{t'}$. 

Although we are interested in the undiscounted objective ($\gamma = 1$), a discount factor $\gamma < 1$ can help obtain better results by further reducing variance at the cost of a small bias \citep{gae, gpomdp}. We are particularly interested in the completely greedy setting - where $\gamma = 0$ - as this leads to particularly simple implementations.

\subsection{MRI subsampling}
We consider a dataset $\mathcal{D}$ of image vectors $\bm{x} \in \mathbb{R}^N$ corresponding to true MR images to be reconstructed from a subsampled k-space signal. The full k-space signal $\bm{y}_N \in \mathbb{C}^N$ is obtained from $\bm{x}$ by a Fourier transform $F \in \mathbb{C}^{N \times N}$ as $\bm{y}_N = F \bm{x}$. The subsampling operator (or mask) $U_m$ that selects $m \leq M < N$ measurements for a total sampling budget $M$ can be represented as an $m \times N$ matrix with every row a one-hot vector of size $N$. Subsampled k-space can then be written $\bm{y}_m = U_m F \bm{x}$. The ratio $N/m$ is called the acceleration of the MR reconstruction process. 

We consider subsampling strategies $\pi(\bm{y}_m )$ that select (possibly stochastically) which k-space measurement should be selected next given the subsampled k-space so far. We would like to find the strategy $\pi$ that yields samples that allow for good reconstructions under reconstruction process $G_\theta$ (where $\theta$ are any parameters in the reconstruction process). The quality of the reconstruction is measured using a quality metric $\eta(\bm{x}, \cdot)$. Formally, we would thus like to find the strategy that maximises the expected quality after $M$ steps:
\begin{equation} \label{eq:mri}
\pi^* = \argmax_\pi \eta(\bm{x}, G_\theta(U_{m+1} F \bm{x} )), \quad U_{m+1} = \begin{bmatrix}
U_{m} \\ k^\mathsf{T}
\end{bmatrix}, \quad k \sim \pi(\bm{y}_{m}).
\end{equation}
In this equation, $k$ is a one-hot vector sampled from the multinomial distribution specified by $\pi$. It is possible to optimise the subsampling and reconstruction method jointly, in which case optimisation is over both $\pi$ and $\theta$. 

While in principle one may obtain a single element (pixel) of the full 2-dimensional k-space $\bm{y}_N$ (corresponding to a single MR image) per measurement, it is more common to obtain a full column \citep{fastmri}. Such Cartesian trajectories - consisting of subsequently measured columns - are typically more efficient due to physical constraints in MR machines. Furthermore, the reduction in their scan time is linear in the number of measurements done. Importantly, implementing optimised Cartesian sequences in practice requires only small modifications to MR software \citep{cs}. As such, in this work we only concern ourselves with Cartesian samplings, and in what follows have redefined $m, M, N$ as referring to the number of columns in the 2-dimensional k-space.

\section{Method} \label{sec:method}
We now connect equations~\eqref{eq:pg} and~\eqref{eq:mri}. In our approach, the subsampling operator $U_M$ is iteratively constructed by sampling from a policy $\pi_\phi(a_t|\hat{\bm{x}}_t)$, which is a conditional probability distribution over discrete actions (measurements). This policy is output by a policy network - parameterised by $\phi$ - that takes as input the current internal state $\hat{\bm{x}}_t$ given by the current reconstruction $G_\theta(U_{t+L} F \bm{x})$. Here $L$ is the initial number of measurements, such that $\hat{\bm{x}}_0$ corresponds to $G_\theta(U_L F \bm{x})$. We will in the following refer to $T=M-L$ - the number of measurements to acquire - as the acquisition horizon.

Starting from an initial subsampling $U_L$, we compute the policy and sample an action $a_0$. This action corresponds to doing a measurement, which we may write as a one-hot vector $k_0$ of size $N$. The subsampling mask for the next acquisition step is then constructed by concatenating this vector in the column direction to the matrix representation of $U_L$ as $U_{L+1} = \begin{bmatrix} U_L^\mathsf{T} \, k_0 \end{bmatrix}^\mathsf{T}$. This operator is applied to the ground truth k-space, and a new reconstruction $\hat{\bm{x}}_1 = G_\theta(U_{L+1} F \bm{x})$ is obtained. A reward is computed using the criterion $\eta$ and the ground truth image $\bm{x}$. The reconstruction is again input to the policy network, from which a policy $\pi_\phi(a_1|\hat{\bm{x}}_1)$ is obtained. This process is repeated until a total of $M$ measurements have been made (including the initial $L$), corresponding to the $T$ steps of equation~\eqref{eq:pg}. Equation~\eqref{eq:mri} may now be rewritten as an iterative optimisation over the policy network parameters $\phi$. In the following we use $G_t$ as shorthand for
$G_\theta \left( \begin{bmatrix} U_{t-1} \\ k_{t-L-1}^\mathsf{T} \end{bmatrix} F \bm{x} \right)$ with $G_L = G_\theta \left( U_L F \bm{x} \right)$, and write $\pi_{\phi}(G_t)$ for the policy obtained for acquisition step $t$ on MR image $\bm{x}$:
\begin{equation} \label{eq:mriiter}
    \hat{\phi} = \argmax_{\phi} \left\{ \mathbb{E}_{\bm{x} \sim \mathcal{D}} \sum_{t=L}^{M-1} \pi_{\phi}(G_t) \left[ \eta(\bm{x}, G_{t+1}) - \eta(\bm{x}, G_{t}) \right] \right\} = \argmax_\phi J(\phi).
\end{equation} 
Here the expected return $J(\phi)$ has been decomposed into separate reward signals over the sequential acquisition steps, by in each step only considering the reconstruction improvement $r^{\bm{x}} (G_{t}, G_{t+1}) = \eta(\bm{x}, G_{t+1}) - \eta(\bm{x}, G_{t})$. The total return is thus given as the total improvement in reconstruction quality over the full acquisition horizon. The MRI subsampling problem has now been formulated as a maximisation of the expected return under a policy given some initial state, and thus we may use gradient ascent on equation~\eqref{eq:pg} to optimise it. For the criterion $\eta$ we use the Structural Similarity Index Measure (SSIM) \cite{ssim}. The SSIM is a differentiable metric that typically corresponds to human evaluations of image quality more closely than alternatives such as Peak Signal-to-Noise Ratio (PSNR) or Mean-Squared Error (MSE) \cite{fastmrires}. Figure~\ref{fig:loop} summarises the iterative acquisition procedure. 

Typically in RL settings a single action is sampled in each time step. However, when multiple actions can be taken in parallel and the rewards for all these actions can be observed, this counterfactual information can be leveraged to construct strong local baselines \citep{wouterbase, wouterwor}. As we have access to the ground truth k-space for all our MR images $\bm{x}$, we may simulate doing multiple acquisitions in parallel, and compute reconstruction improvements for all of them. Sampling $q$ actions at every time step, and writing $r_{i,t}^{\bm{x}}$ for the reward obtained from sample $i$ at time step $t$, we obtain the following estimators:
\begin{align}
    \nabla_\phi J(\phi) &\approx \frac{1}{q-1} \mathbb{E}_{\bm{x} \sim \mathcal{D}} \sum_{i=1}^q \sum_{t=L}^{M-1} \left[ \nabla_\phi \log \pi_\phi(G_t) \sum_{t'=t}^{M-1} \gamma^{t'-t} \left(r_{i,t'} - \frac{1}{q} \sum_{j=1}^q r_{j,t'} \right) \right] \text{(Non-greedy)}, \label{eq:ngreedy} \\
    \nabla_\phi J(\phi) &\approx \frac{1}{q-1} \mathbb{E}_{\bm{x} \sim \mathcal{D}} \sum_{i=1}^q \sum_{t=L}^{M-1} \left[ \nabla_\phi \log \pi_\phi(G_t) \left(r_{i,t} - \frac{1}{q} \sum_{j=1}^q r_{j,t} \right) \right] \text{(Greedy)}. \label{eq:greedy}
\end{align}
The process for obtaining parallel samples varies between the greedy an non-greedy settings. In the greedy setting, one may easily obtain parallel reward samples by at every time step acquiring $q$ k-space columns in parallel, obtaining the corresponding $q$ reconstructions, and computing the resulting 
\begin{wrapfigure}{r}{0.5\linewidth}
  \centering
  \includegraphics[clip, width=\linewidth]{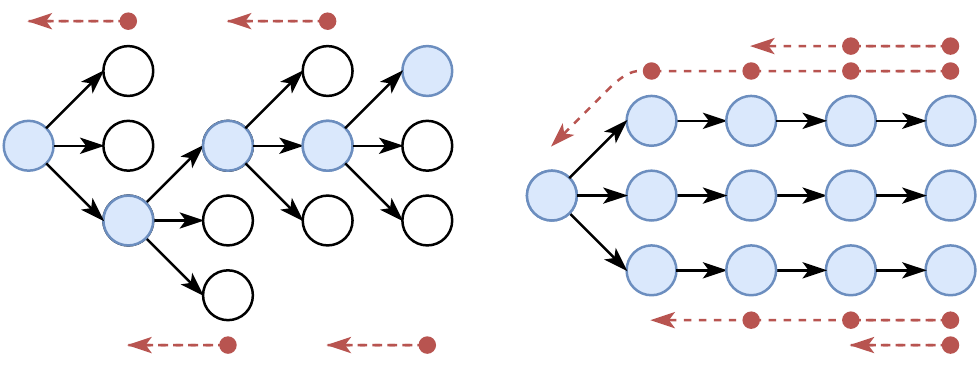}
  \caption{Parallel sampling procedure for the greedy (left) and non-greedy (right) estimators, for an acquisition horizon of four with $q=3$ samples. Open circles represent states, with blue open circles the states selected to continue a trajectory. Solid black arrows represent actions, dashed red arrows represent gradients flowing from rewards in future states of the acquisition horizon to earlier states. Solid red circles on the gradient arrows denote future rewards included in the gradient.}
  \label{fig:psamp}
\end{wrapfigure}
rewards. Since the objective~\eqref{eq:greedy} relies only on immediate rewards, gradients can be immediately computed. In the non-greedy setting~\eqref{eq:ngreedy} however, entire trajectories must be sampled before a single optimisation step may be performed. Sampling multiple actions in every state along the trajectory leads to a combinatorial explosion in number of rewards that need to be computed and gradients that need to be stored. To circumvent this issue, in the non-greedy setting we sample multiple actions in the initial state only. This initialises parallel trajectories, which may be traversed using a single sample at every time step (as in more typical reinforcement learning settings). Local baselines for a given time step and trajectory are then computed using the reward corresponding to that time step in the parallel trajectories. Figure~\ref{fig:psamp} illustrates the parallel sampling procedure and associated gradient computation.


\section{Experiments}
\subsection{Implementation}
\paragraph{Datasets:} We leverage the NYU fastMRI open database containing a large number of knee and brain volumes for our experiments \citep{fastmri}, restricting ourselves to single-coil scans. The provided test volumes do not contain full ground-truth k-space measurements, so we split 20\% of the training data into a test set. For the knee data, we use half of the available volumes for computational expedience, and additionally remove the outer slices of each volume as in \citep{irismri}, since those typically depict background. This leads to a dataset of 6959 train slices, 1779 validation slices, and 1715 test slices. The slices were cropped to the central $128\times128$ region as in \citep{jin19}, again to save on computation. We will refer to this dataset as the Knee dataset from now on. 

We also construct a Brain dataset from the fastMRI brain volumes. Here the full singlecoil k-space is simulated from the ground truth images by Fourier transform after cropping to the central $256\times256$ region. This allows us to test whether policy gradient models can scale to images larger than $128\times128$. We use a fifth of the available volumes for computational reasons, resulting in 11312 train slices, 4372 validation slices, and 2832 test slices.

\paragraph{Reconstruction model:} Jointly training our policy networks with a reconstruction model - while possible - adds variance to the training process: this has effects on the policy models that are hard to predict. Since we are primarily interested in analysing the subsampling strategy, we instead opt to use a pretrained reconstruction model in our experiments. Many powerful reconstruction networks are already available \citep{fastmrires}, motivating the need for subsampling methods that may be trained on top of any of these models. Additionally, the use of a fixed, pretrained reconstruction model allows us to show that adaptivity indeed improves performance of subsampling methods, by comparing to a non-adaptive oracle that requires a predefined reconstruction method to make predictions. This intuitive notion has, to our knowledge, not yet been empirically quantified. For the reconstruction model we use the standard 16-channel U-Net baseline provided in the fastMRI repository. Hyperparameters and training details are left mostly unchanged: we refer to Appendix~\ref{sec:apprec} for further details.

\paragraph{Policy models:} We train greedy and non-greedy policy models using equations~\eqref{eq:greedy} and~\eqref{eq:ngreedy} respectively. Setting $\gamma = 1.0$ in equation~\eqref{eq:ngreedy} corresponds to the undiscounted objective, where the policy gradient is an unbiased estimate of the gradient of the expected return: we refer to this as the (fully) non-greedy model. We additionally report results on $\gamma = 0.9$, as this model performed best in our experiments (see Appendix~\ref{sec:appssim} for additional results). 

Two initial accelerations are considered: subsampling by a factor 8 and 32. For Knee data, respectively 16 and 28 acquisitions are performed, resulting in a final acceleration of 4 in both cases. We refer to these two settings as respectively the `base' and `long' horizon settings. For Brain data we perform the same procedure, but note that here the final mask does not correspond to an acceleration of 4, due to the larger image size. Reconstructions are initialised by obtaining low frequency (center) columns of k-space equal to the initialisation budget. The process as described in section~\ref{sec:method} is then performed, with $q=8$ for both estimators. Models are trained for 50 epochs using a batch size of 16. We use the same architecture for the greedy and non-greedy policy models, as experimentation with larger and smaller architectures showed no clear improvements. Further model details are contained in Appendix~\ref{sec:apppolarch} and~\ref{sec:apppolhyp}, as well as a comparison of computational load in~\ref{sec:appcomp}. The greedy model is somewhat simpler to implement and significantly less costly to train than the non-greedy model.

\paragraph{AlphaZero model:} \label{sec:alpha} The non-greedy, adaptive, AlphaZero-inspired method of \citep{jin19} provides a literature baseline for comparison with our models. Their model can be adapted to our task, as it is flexible enough to learn a policy given any reconstruction method and reward signal.\footnote{We thank the authors for access to their research code and their advice in selecting appropriate hyperparameters. The long horizon setting is inspired by their suggestions.} 

To make a proper comparison, we use the output of our fixed reconstruction model as input to their policy model. The policy model is trained in a self-supervised manner by means of Monte Carlo tree search, using SSIM for the reward signal rather than the original PSNR. In the following, we will refer to this model as AlphaZero. Due to computational constraints, we were only able to train this model on the Knee dataset. Further details can be found in Appendix~\ref{sec:appalph}.

\subsection{Results} \label{sec:res}
In Table~\ref{tab:ssims} we report average and standard deviation test data SSIM scores of the final reconstruction obtained under various models after exhausting the sampling budget. `Greedy' and `NGreedy' are our proposed models, and `$\gamma = 0.9$' is the non-greedy model with discount factor $\gamma = 0.9$ instead of $\gamma = 1.0$. `AlphaZero' refers to the model of \citep{jin19} adapted to our task as described in section~\ref{sec:alpha}. Also compared is a non-adaptive oracle, denoted as `NA Oracle'. This oracle selects as a measurement candidate in each acquisition step the column that leads to the greatest average SSIM improvement over the test dataset. This is not a feasible strategy without direct access to the ground truth k-space, and provides an upper bound on the performance of greedy models that do not adapt their predictions to individual MR slices. Additionally, we compare a `Random' strategy that shares the initial mask with the other methods, and subsequently obtains a uniformly random measurement every acquisition step, similar to a simple VDS heuristic. See Appendix~\ref{sec:appssimsup} for extended results.
\begin{table}
    \caption{SSIM performance on test data. For non-deterministic models, averages and standard deviations are computed over five seeds, using $q=8$ trajectories for policy models (AlphaZero scores are averaged over three seeds instead).}
    \label{tab:ssims}
    \centering
    \begin{tabular}{lcccc}
        \toprule
         & \multicolumn{2}{c}{\textbf{Knee}} & \multicolumn{2}{c}{\textbf{Brain}} \\
         \cmidrule(r){2-3} \cmidrule(r){4-5}
         & \textbf{Base horizon} & \textbf{Long horizon} & \textbf{Base horizon} & \textbf{Long horizon} \\
         \midrule
         \textbf{Random} & $0.6948\!\pm\!0.0003$ & $0.6602\!\pm\!0.0006$ & $0.9020\!\pm\!0.0001$ & $0.5820\!\pm\!0.0006$ \\
         \textbf{NA Oracle} & $0.7213$ & $0.7421$ & $0.9099$ & $0.8909$ \\
         \textbf{AlphaZero} & $0.7203 \pm 0.0008$ & $0.7403 \pm 0.0009$ & - & - \\
         \textbf{NGreedy} & $0.7223 \pm 0.0003$ & $0.7421 \pm 0.0014$ & $0.9103 \pm 0.0002$ & $0.8886 \pm 0.0048$ \\
         \textbf{Greedy} & $0.7230 \pm 0.0001$ & $0.7442 \pm 0.0007$ & $0.9106 \pm 0.0001$ & $0.8917 \pm 0.0002$ \\
         $\bm{\gamma = 0.9}$ & $0.7232 \pm 0.0002$ & $0.7449 \pm 0.0004$ & $0.9106 \pm 0.0003$ & $0.8921 \pm 0.0001$ \\
        \bottomrule
    \end{tabular}
\end{table}

Our Greedy model obtains SSIM scores superior to or on-par with most compared models on all tasks. Of the methods reported here, only the $\gamma = 0.9$ model ostensibly outperforms it, although the differences are (close to) within one standard deviation. The Greedy model is furthermore much less computationally expensive to train than any of the non-greedy models. In the following we focus our attention on the Greedy and NGreedy models specifically, as these two extrema prove illustrative for our analysis of adaptivity.

\paragraph{Adaptivity:} The Greedy model shows in Table~\ref{tab:ssims} that it is adapting its predictions to individual MR images by outperforming the NA Oracle. This suggests that adaptivity is indeed a useful property for subsampling models to possess, and that this information can be learned in practice. Our NGreedy model outperforms the NA Oracle only clearly on the base horizon task, suggesting that for longer horizons it may fail to be sufficiently adaptive. Note that the NGreedy model has two avenues for potentially improving over the NA Oracle: adaptivity, and non-greediness. 

\begin{figure}
  \centering
  \includegraphics[width=\linewidth]{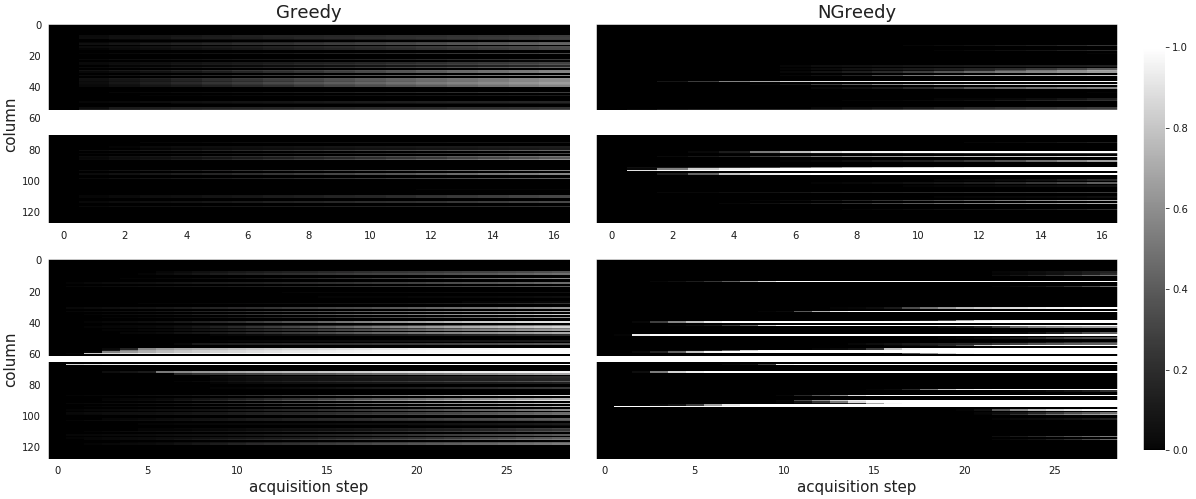}
  \caption{Visualisation of sampled trajectories for base (top) and long (bottom) horizons, averaged over the Knee test data. Shown is the fraction of MR slices for which a particular column has been sampled at an acquisition step. The central white bands are initialisation measurements. The NGreedy policies select the same column for more slices than the Greedy model, suggesting it adapts less to the individual images. For each setting the best model on the test set was used.}
  \label{fig:pol}
\end{figure}

To further investigate this behaviour, we visualise the average (over all slices in the Knee dataset) of the learned policies at every acquisition step in Figure~\ref{fig:pol}. Notable is that the NGreedy model outputs more sharply peaked average policies than the Greedy model, suggesting a lack of adaptivity to the input (see Appendix~\ref{sec:apppolvis} for more average policy visualisations).
\paragraph{Mutual information:} As adaptivity is required for a greedy model to outperform the NA Oracle, we may conclude that the Greedy policies are adaptive, but it would be prudent to quantify this effect. The mutual information (MI) can be used as a quantitative measure of how much information an observed state (reconstruction) gives about the action (measurement location) under the learned policy. This mutual information provides a direct measure of how adaptive a model is: the higher the MI, the more the model changes its policy as the state changes. For details see Appendix~\ref{sec:appmimot}. 

In Figure~\ref{fig:mikneegnggamma09} we visualise the MI per acquisition step for the Greedy, NGreedy and $\gamma = 0.9$ models on Knee data. 
\begin{figure}
  \centering
  \includegraphics[width=\linewidth]{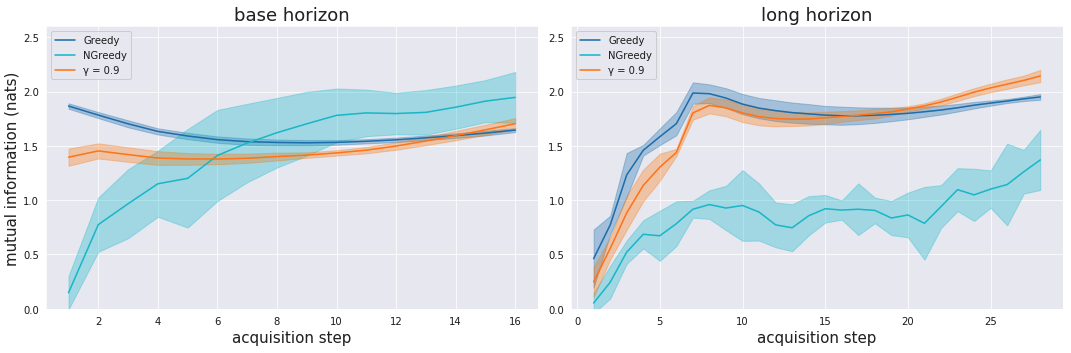}
  \caption{Mutual information for the base (left) and long (right) horizon settings for the Greedy, NGreedy and $\gamma = 0.9$ methods on Knee test data. Shown is the average and standard deviation of the mutual information per acquisition step over five seeds, computed with $q=8$ trajectories.}
  \label{fig:mikneegnggamma09}
\end{figure}
For the base horizon task, it seems the Greedy model learns to be adaptive already in the first acquisition, having enough information to produce adaptive policies. The NGreedy model has low mutual information for the initial acquisition steps, but becomes more adaptive as it gets closer to exhausting the acquisition budget. For the long horizon task the differences in MI are much starker, which is consistent with our observation that the NGreedy model fails to achieve the adaptivity required to outperform the NA Oracle. Note that $\gamma = 0.9$ seems to interpolate between Greedy and NGreedy, and behaves more similar to former, even though its objective and training details are more alike the latter. Similar results are obtained for the Brain dataset. We refer to Appendix~\ref{sec:appmi} for a more comprehensive analysis of the MI.

\paragraph{Signal-to-Noise:} We hypothesise the lack of adaptivity - of the NGreedy model relative to the Greedy - is partially due to higher variance gradient estimates in the NGreedy estimator, resulting from the longer acquisition horizon. Rather than learning to adapt the policy to individual MR slices, the NGreedy models seems to use the average reward signal to learn policies that perform well on average. The signal-to-noise ratio (SNR) provides a quantitative measure that can be used to compare the variance of our gradient estimators \citep{snrpg}. Taking inspiration from \citep{snrmax}, we compute the SNR of a gradient estimate as the empirical mean gradient $\hat{\mu}$ divided by the empirical standard deviation $\hat{\sigma}_\mu$ of this mean gradient. The mean gradient is computed over training data as the average of gradients $g_i$ obtained from batch $i \in [1, B]$: $\hat{\mu} = \frac{1}{B} \sum_{i=1}^B g_i$. For large enough $B$, the variance of $\hat{\mu}$ is well estimated as $\hat{\sigma}^2_\mu = \frac{1}{B(B-1)} \sum_{i=1}^B (g_i - \hat{\mu})^2$ due to the Central Limit Theorem \citep{snrmax}. 

Estimated SNR on the training data is depicted in Table~\ref{tab:snr}. SNR is computed over the gradients of the weights in the final layer of the policy network. The non-greedy models consistently have lower SNR than the Greedy model for both horizons and datasets, consistent with our hypothesis. Additionally, the lower SNR for NGreedy estimators on the long versus the base horizon task suggests that higher variance is associated with longer horizons, although we note that this effect does not hold for $\gamma = 0.9$ models. This analysis suggests that the non-greedy models obtain lower bias in the optimisation objective at the cost of higher variance, as was already observed in ~\citep{gae}. For these specific experiments, the sweet spot in this trade-off seems to lie around $\gamma = 0.9$ (see Appendices~\ref{sec:appssimsup} and~\ref{sec:appsnr}).
\begin{table}
    \caption{Signal-to-Noise ratio comparison of the Greedy and NGreedy models for the two time horizons trained on. Displayed are average SNR estimates and standard deviations obtained for the best performing model on three runs over the test data for every setting ($q=16$ samples, batch size $16$). Epoch $n$ refers to the model after the $n$'th epoch of training has been completed.}
    \label{tab:snr}
    \centering
    \begin{tabular}{lcccccc}
        \toprule
         & \multicolumn{6}{c}{\textbf{Knee}} \\
         & \multicolumn{3}{c}{\textbf{Base horizon}} & \multicolumn{3}{c}{\textbf{Long horizon}} \\
         \cmidrule(r){2-4} \cmidrule(r){5-7}
         & \textbf{Greedy} & \textbf{NGreedy} & $\bm{\gamma = 0.9}$ & \textbf{Greedy} & \textbf{NGreedy} & $\bm{\gamma = 0.9}$ \\
         \midrule
        \textbf{Epoch 1} & $2.21\!\pm\!0.24$ & $1.82\!\pm\!0.01$ & $2.22\!\pm\!0.18$ & $2.46\!\pm\!0.25$ & $1.68\!\pm\!0.14$ & $2.05\!\pm\!0.11$ \\
        \textbf{Epoch 20} & $3.91\!\pm\!0.08$ & $1.24\!\pm\!0.07$ & $2.12\!\pm\!0.06$ & $3.49\!\pm\!0.27$ & $1.04\!\pm\!0.16$ & $3.04\!\pm\!0.06$ \\
        \textbf{Epoch 50} & $2.51\!\pm\!0.03$ & $1.02\!\pm\!0.07$ & $1.43\!\pm\!0.10$ & $2.15\!\pm\!0.16$ & $0.96\!\pm\!0.16$ & $1.29\!\pm\!0.11$ \\
        \midrule
         & \multicolumn{6}{c}{\textbf{Brain}} \\
         & \multicolumn{3}{c}{\textbf{Base horizon}} & \multicolumn{3}{c}{\textbf{Long horizon}} \\
         \cmidrule(r){2-4} \cmidrule(r){5-7}
         & \textbf{Greedy} & \textbf{NGreedy} & $\bm{\gamma = 0.9}$ & \textbf{Greedy} & \textbf{NGreedy} & $\bm{\gamma = 0.9}$ \\
         \midrule
        \textbf{Epoch 1} & $6.70\!\pm\!0.09$ & $3.76\!\pm\!0.22$ & $5.31\!\pm\!0.10$ & $8.75\!\pm\!0.20$ & $1.57\!\pm\!0.11$ & $7.80\!\pm\!0.10$ \\
        \textbf{Epoch 20} & $11.21\!\pm\!0.08$ & $2.95\!\pm\!0.23$ & $5.32\!\pm\!0.02$ & $13.36\!\pm\!0.19$ & $1.18\!\pm\!0.15$ & $7.22\!\pm\!0.19$ \\
        \textbf{Epoch 50} & $7.02\!\pm\!0.07$ & $1.45\!\pm\!0.10$ & $2.35\!\pm\!0.00$ & $4.56\!\pm\!0.09$ & $0.82\!\pm\!0.08$ & $2.98\!\pm\!0.09$ \\
        \bottomrule
    \end{tabular}
\end{table}

\paragraph{AlphaZero:} The AlphaZero method of \citep{jin19} slightly underperforms our NGreedy model. Although not a policy gradient approach, we expect it experiences some of the same optimisation problems: high variance in the gradients due to optimisation over long acquisition horizons. However, due to computational constraints (these models take up to a week to train) we were not able to do an extensive hyperparameter search, so it may be possible to further improve this model's performance.

\section{Conclusion and discussion}
We have proposed a practical, easy to train greedy model that learns MRI subsampling policies on top of any reconstruction method, using policy gradient methods. In our experiments this greedy model performs nearly on-par with the most performant model tested, is the simplest to implement, and requires fewest computational resources to train. We have furthermore observed that fully non-greedy models perform worse than their greedier counterparts, and hypothesised that this is due to the former primarily learning from an average reward signal, due to high variance in the optimisation. We have provided a number of experiments to support this hypothesis, and have moreover shown adaptivity to individual MR slices to be an useful property in a sampling policy. To our knowledge, this is the first time the MRI subsampling problem has been analysed on these axes in a deep learning context. Our results suggest that future methods for learning subsampling masks in MRI might profit from greedier / discounted optimisation objectives. Additionally, such methods might do well to incorporate adaptivity in their mask designs.

Our analysis was performed on single-coil volumes. We suspect our conclusions will hold in the more clinically relevant multi-coil settings, but this is to be confirmed by future work \citep{fastmri}. The reward baseline used for the greedy model cannot be used for the non-greedy model due to a combinatorial explosion in the required number of sampled trajectories. Future research may investigate incorporating value function learning~\citep{sutton} to estimate the return of a particular node in Figure~\ref{fig:psamp}, thus enabling the more local reward baseline to be used for the non-greedy setting as well. Interestingly, for certain types of subsampling problems, theoretical bounds exist that indicate that optimal adaptive greedy strategies perform almost as well as their optimal non-greedy counterparts \citep{adapsub, fujii}. Investigating to what degree such bounds hold for the MR subsampling problem is a promising topic for future work as well.

\section*{Broader Impact}
Accelerating MRI reconstructions beyond the current standard has the potential to improve patient satisfaction and throughput. In this paper we have attempted to analyse the subproblem of designing optimal sampling strategies with deep learning methods in a way that provides future research with more principled motivations for choosing a particular approach to this problem. However, more work is needed before the insights of this paper become directly clinically relevant. In particular, it is unclear that simple image quality metrics (such as the SSIM used in this work) provide an adequate measure of clinical usability \citep{cheng}. While the results of the fastMRI reconstruction challenge \citep{fastmrires} suggest that the SSIM does provide estimates of image quality consistent with the preferences of radiologists, it is also noted that current reconstruction methods have trouble identifying subtle pathologies, as these are smoothed out by the average signal from the dataset. The authors of \citep{fastmrires} suggest that more work is needed relating radiologists' preferences to image quality metrics at the level of diagnostic interpretation. Until then, methods optimised on such metrics should be used in clinical settings only with extreme care, as they risk falsely declaring a patient free of potential health- and life-threatening pathologies.

\section*{Acknowledgements and disclosure of funding}
We thank Marco Federici, Shi Hu, Maximilian Ilse, Daniel Worrall, and especially Wouter Kool and Bas Veeling for useful discussions. We are grateful to the Weights\&Biases team \citep{wandb} for providing their experiment tracking software. We would further like to thank the three anonymous NeurIPS2020 reviewers for their detailed and helpful feedback.

This work is supported by the ‘Efficient Deep Learning’ (EDL, \url{https://efficientdeeplearning.nl}) research programme, which is financed by the Dutch Research Council (NWO) domain Applied and Engineering Sciences (TTW). 

We furthermore would like to disclose that M. Welling is Vice President of Technologies at Qualcomm Technologies Netherlands, in addition to his university position.

\bibliography{references.bib}

\clearpage

\appendix

\section{Implementation details}

\subsection{Reconstruction model} \label{sec:apprec}
As stated in the main text, our reconstruction model is the standard 16-channel U-Net baseline provided in the fastMRI repository pulled in November 2019.\footnote{https://github.com/facebookresearch/fastMRI/tree/a55a1b129eb1d98ec9df26bfa2617a3b8c957d21} Hyperparameters are left unchanged, except for a switch of optimiser from RMSProp to Adam \citep{adam}, and training for the full 50 epochs rather than doing early stopping based on the validation set. 

The input to the model consists of (real-valued) so-called \textit{zero-filled} images, obtained by applying the inverse Fourier transform to subsampled k-space and taking the complex norm of the resulting image. The full k-space is obtained from the ground truth images by Fourier transform after cropping to $(128 \times 128)$ pixels for Knee data, and $(256 \times 256)$ pixels for Brain data. Note that we crop in image space - rather than k-space - which reduces computation while preserving image detail. We train on accelerations $(4, 4, 4, 6, 6, 8)$, with center fractions $(0.25, 0.167, 0.125, 0.167, 0.125, 0.125)$. This means masks contain a low-frequency (central) k-space region that is always sampled, and have up to half of the budget randomly sampled in the remainder of k-space. The random sampling is done so that the reconstruction model learns to reconstruct for a wide variety of masks \citep{fastmri}. In contrast to our policy models, we only have to train the reconstruction model once for each data set, and so we train on all the available training volumes. We use these models pretrained in our further pipeline. The model has 837,635 parameters.

\subsection{Policy model architecture} \label{sec:apppolarch}
We use slightly different policy model architecture for the Knee and Brain model, primarily to keep the number of parameters similar. No changes are made to the model architecture when switching between the Base and Long horizon setting.

\subsubsection{Knee model architecture}
Starting from the $(128\times128)$ reconstructed image, an initial $(1\times1)$ convolution is applied to upsample to 16 channels. We follow this by instance normalisation and ReLU activation. We further employ four convolutional blocks, each consisting of a zero-padded $(3\times3)$ convolution layer that doubles the number of channels, followed by an instance normalisation, ReLU activation, and $(2\times2)$ max-pooling layer.

The resulting $(8\times8\times256)$ tensor is flattened and fed through a dense layer of 256 neurons, followed by a leaky-ReLU activation with slope $0.01$. This is followed by another such layer and activation, before a final dense layer with 128 neurons and a Softmax operation to turn the output into probabilities corresponding to the columns in k-space. The model has 4,685,568 parameters.

As mentioned in the main text, reconstructions are initialised by obtaining low-frequency (center) columns of k-space equal to the initialisation budget. This corresponds to 16 columns for acceleration 8 and to 4 columns for acceleration 32. After initialisation, the process described in section~\ref{sec:method} is performed, setting $q=8$ samples for both estimators. For the size 128 images used in this work, this process corresponds to respectively 16 and 28 acquisition steps, ending at an acceleration factor of 4 for both settings. Models are trained for 50 epochs using a batch size of 16. A single gradient step is performed after accumulating gradients for a full acquisition trajectory. 

\subsubsection{Brain model architecture}
The differences between the Brain and Knee policy model architectures are slight: the Brain model uses five convolutional blocks of the type described rather than four, and the initial upsampling is to 8 channels, rather than 16. Because the Brain model input is of size the $(256\times256)$, these choices ensure the feature representation after the final convolutional block has size $(8\times8\times256)$ as well. The model has 4,719,552 parameters.

As with the Knee model, we train the Brain model starting with accelerations 8 and 32, acquiring a further 16 and 28 k-space columns with our policy, for the Base and Long horizon cases respectively. Note that since the Brain images are larger, the final state does not correspond to the acceleration factor 4 of the Knee setting. Instead, we end up with $\frac{256}{8} + 16 = 48$ and $\frac{256}{32} + 28 = 36$ columns for the Base and Long horizon settings respectively, corresponding to acceleration factors of $\frac{256}{48} = 5\frac{1}{3}$ and $\frac{256}{36} = 7\frac{1}{9}$. This choice was made due to computational concern, as training the Brain models takes up to three times as long as training Knee models, and requires more RAM (see section~\ref{sec:appcomp}).

\subsection{Policy model hyperparameters} \label{sec:apppolhyp}
Hyperparameter tuning with random search on Knee data found little performance differences using larger models or longer training. The most influential hyperparameters proved to be learning rate, batch size, and number of samples per acquisition step. Values of the latter two are constrained by memory considerations during training of non-greedy models (see section~\ref{sec:appcomp}), and were set to their highest reasonable values of 16 and 8 respectively. Learning rate was further tuned by hand for both Knee models individually. We did not do any additional hyperparameter tuning for the Brain models, opting to use the exact same settings as we used for the Knee models.

The Greedy and NGreedy models are both trained with a learning rate of $5\mathrm{e}{-5}$. The learning rate is decayed once by a factor 10 after 40 epochs for the Greedy model, and decayed a factor 2 every 10 epochs for the NGreedy model, for a total decay rate of 16. Training was done using the Adam optimiser with no weight decay. The $\gamma = 0.9$ model is trained with the same parameter settings as the NGreedy model. Individual test scores of the runs presented in Table~\ref{tab:ssims} of the main text were computed by averaging scores of 8 trajectories.

As it is always known which k-space columns have already been measured, we artificially set the probabilities for these columns to 0 during training and evaluation. This ensures the model is focused on the task of finding the optimal policy, rather than also trying to learn which measurement have already been done given only the reconstructed input image to go on. We have experimented with instead feeding this as extra information to the policy model, but this tended to destabilise training.

\subsection{AlphaZero hyperparameters} \label{sec:appalph}
The AlphaZero model architecture used is as defined in Figure 6 of \citep{jin19}, with 32 channels rather than the original 64, as we saw no performance differences in initial experiments. It was implemented using the research code provided to us in private communication. Training used the original 540 rounds, but model training was early-stopped when the final SSIM (after the last acquisition step) on the validation set flattened out. Due to computational constraints, we were unable to do a proper hyperparameter search, and as such have left the remaining hyperparameters unchanged from their default values: brief experimentation with slightly changed hyperparameter settings showed no obvious performance differences.

Individual test scores of the runs presented in Table~\ref{tab:ssims} of the main text were computed by averaging scores of 8 trajectories, the same as for the Greedy and NGreedy policy gradient models. The model has 27,558,849 parameters, about five to six times more than our policy models.

\subsection{Policy gradient model pseudocode}
In Algorithm~\ref{alg:train} we provide pseudocode for a training epoch of our policy gradient models. As illustrated by Figure~\ref{fig:psamp} of the main text, the greedy and non-greedy estimators~\eqref{eq:greedy} and~\eqref{eq:ngreedy} compute different reward baselines. In the pseudocode this is controlled by the $IsGreedy$ argument.

\begin{tcolorbox}[fonttitle=\bfseries, title=Policy model train epoch]
\begin{algorithm}[H] \label{alg:train}
\DontPrintSemicolon
\caption{Pseudocode for a train epoch of the policy gradient models.}
Algorithm parameters: number of trajectories $q$, discount factor $\gamma$, IsGreedy True/False, number of acquisition steps $T$\;
Initialise reconstruction model $G$, initial mask $U_0$, train dataset $\mathcal{D}$ containing batches of ground truth MR images $\bm{x}$, metric $\eta$ (e.g. SSIM)\;
\ForEach{\text{batch} in $\mathcal{D}$}{
    Compute initial reconstructions $g_0 \leftarrow G(U_0 F\bm{x})$\;
    Compute initial metric $v_0 \leftarrow \eta(g_0, \bm{x})$\; 
    \For{$t \in \{1, ..., T\}$}{
        Compute policy $\pi(\cdot|g_{t-1})$ given the current reconstruction\;
        Sample $q$ actions $a_t$ from $\pi(\cdot|g_{t-1})$\;
        Obtain probabilities $p_t \leftarrow \pi(a_t|g_{t-1})$\;
        \eIf{IsGreedy}{
            Append these actions to $q$ copies of $U_{t-1}$ to form $U_t$\;
        }{
            \eIf{$t=1$}{
                Append these actions to $q$ copies of $U_{t-1}$ to form $U_t$\;
            }{ 
                Append these $q$ actions to the $q$ instances of $U_{t-1}$ in memory to form $U_t$\;
            }
        }
        Compute next-step reconstructions $g_t \leftarrow G(U_t F\bm{x})$\;
        Compute metrics $v_t \leftarrow \eta(g_t, \bm{x})$\;
        Compute rewards $r_t \leftarrow v_t - v_{t-1}$\;
        Store log probabilities $\log(p_t)$ of actions, and rewards $r_t$\;
        \eIf{IsGreedy}{
            Compute loss according to Equation~~\eqref{eq:greedy}\;
            Store gradient updates (e.g. loss.backward() in PyTorch) \;
            Randomly select one of the $q$ copies of $U_t$ to continue with\;
        }{
            \If {$t = T$}{
                Compute loss according to Equation~\eqref{eq:ngreedy}\;
                Store gradient updates\;
            }  
            Continue with all $q$ instances of $U_t$\;
        }
    Update policy model weights using stored gradient updates (e.g. optimizer.step() in PyTorch) \;
    }
}
\end{algorithm}
\end{tcolorbox}

\subsection{Lack of comparison to Zhang \textit{et al.} (2019)} \label{sec:zhang}
Ideally, we would wish to compare our method to the greedy acquisition method of \citep{zhang}. Unfortunately, there are a number of reasons that make this infeasible, which we discuss here.

As mentioned in the main text (Section~\ref{sec:rel}), the approach in \citep{zhang} requires joint training of the reconstruction network with an evaluator network that guides acquisition through a similarity score between ground truth and fantasised k-space. Joint training is crucial, as the reconstruction network must be incentivised to produce reconstructions that have consistent k-space representation for evaluator based acquisition to perform well. This contrasts with our method, where joint training is optional, and our acquisition function is directly (reinforcement) learned using policy gradients on image-space input. This also poses a challenge for making a fair comparison (using the same reconstruction model): the reconstruction model in \citep{zhang} is incentivised to care about features that are not necessarily relevant to our policy, and our reconstruction method is not necessarily incentivised to care about features that are crucial to their evaluator. 

To explore these differences we performed a proxy comparison using our reconstruction model and replacing their evaluator score with the true spectral map score computed from ground truth images. Using ground truth test images makes this an oracle method - infeasible in practice - but provides an upper bound for the performance of \citep{zhang} under our reconstruction model, as we now use true spectral map scores, rather than the estimate learned by the evaluator network. However, this oracle method performed far worse than our models, suggesting that the strategy in \citep{zhang} indeed depends heavily on reconstruction model design choices that force consistency of k-space, as well as on joint training with the evaluator. We also note that there is no code available for \citep{zhang}, further complicating attempts at a fair comparison.

\subsection{Dynamic range and SSIM}
SSIM hyperparameters are kept to their original values in \citep{ssim}. The dynamic range is a dataset dependent hyperparameter of the SSIM metric that encodes the value range of a particular image. For an MRI slice the dynamic range is typically chosen to be maximum pixel value in the corresponding ground truth volume \citep{fastmri}.

\subsection{PSNR evaluation}
We have used the Structural Similarity Index Measure (SSIM) \citep{ssim} as a reward signal in this work. The SSIM is a differentiable metric that typically corresponds to human evaluations of image quality more closely than common alternatives \citep{fastmrires}. One such alternative is the Peak Signal-to-Noise Ratio (PSNR). While we do not train on this metric, it may still be insightful to evaluate the trained models by it.

Interestingly however, evaluating the SSIM non-adaptive and adaptive oracles on Knee data with PSNR, results in scores of 27.21, 27.50 on the base horizon, and 25.59, 26.13 on the long horizon task respectively. In contrast, SSIM scores were clearly higher for the Knee data long horizon task, and indeed this is what one would expect for oracles. This suggests that SSIM and PSNR care about distinct features (notably, PSNR seems to favour more low-frequency columns), which complicates drawing conclusions from PSNR evaluations of SSIM optimised methods: verification of reconstruction quality by human experts seems necessary in order to draw further conclusions. Because of this complication, we have opted not to report PSNR scores, though note that the provided code does provide PSNR evaluations.

\section{Further analysis}

\subsection{Comparison of computational load} \label{sec:appcomp}
As the architecture for the Greedy and NGreedy models is equivalent, differences in computational load required for training the models stem from variations due to the different optimisation objectives. As the Greedy model requires only looking ahead a single acquisition step, we are only required to store the cumulative gradient of the full acquisition trajectory. The NGreedy model however requires storing individual gradient information for all parallel acquisition trajectories and all acquisition steps, as the full return is only known after performing the last measurement of the budget. 

These memory requirements constrain the batch size significantly for the NGreedy model - approximately linearly in length of the optimisation horizon - adding to gradient variance. To circumvent this issue and make a proper comparison with the Greedy model, we accumulate gradients for multiple batches, doing an optimisation step only when the effective batch size has reached that of the Greedy model (dividing the accumulated gradients by the number of batches used to accumulate them to properly mimic training over larger batches). However, this procedure effectively increases the number of batches treated sequentially by a factor equal to the ratio in batch size, slowing down training correspondingly.

We note that the memory burden of the NGreedy model may also be heavily reduced by storing (state, action, reward)-transition tuples for every trajectory encountered during an episode, and discarding model gradients at this step. Then, when the episode has concluded, the full return as well as the log probabilities of the stored action may be computed and the gradient backpropagated as normal. This requires one more forward pass of the policy model for every action (to compute the log probability gradients), but frees up storage space because these gradients need not be remembered for the full trajectory. Especially for longer horizons this may speed up computation by exploiting the use of larger batches than otherwise possible.

We refer to Table~\ref{tab:comp} for a comparison of (very) approximate training times on GTX1080Ti GPUs. These numbers are based on observation of the training logs, not on an exact computation, as the latter was rendered impossible due to issues on the GPU clusters that caused training jobs to crash halfway through. The largest source of variance in training time within a setting seems to be related to I/O, likely due to the loading of large MR image files.
\begin{table}
    \caption{Approximate training times in days for policy models on GTX1080Ti GPUs. Number of GPUs used in parallel given in brackets.}
    \label{tab:comp}
    \centering
    \begin{tabular}{lcccc}
        \toprule
         & \multicolumn{2}{c}{\textbf{Base horizon}} & \multicolumn{2}{c}{\textbf{Long horizon}} \\
         \cmidrule(r){2-3} \cmidrule(r){4-5}
         & \textbf{Greedy} & \textbf{NGreedy} & \textbf{Greedy} & \textbf{NGreedy} \\
         \midrule
        \textbf{Knee} & 1.0 (1) & 1.5 (1) & 1.5 (1) & 2.0 (2) \\
        \textbf{Brain} & 2.0 (1) & 3.0 (2) & 4.5 (1) & 5.0 (4) \\
        \bottomrule
    \end{tabular}
\end{table}


\subsection{Additional policy visualisations} \label{sec:apppolvis}
\begin{figure}
  \centering
  \includegraphics[width=.8\linewidth]{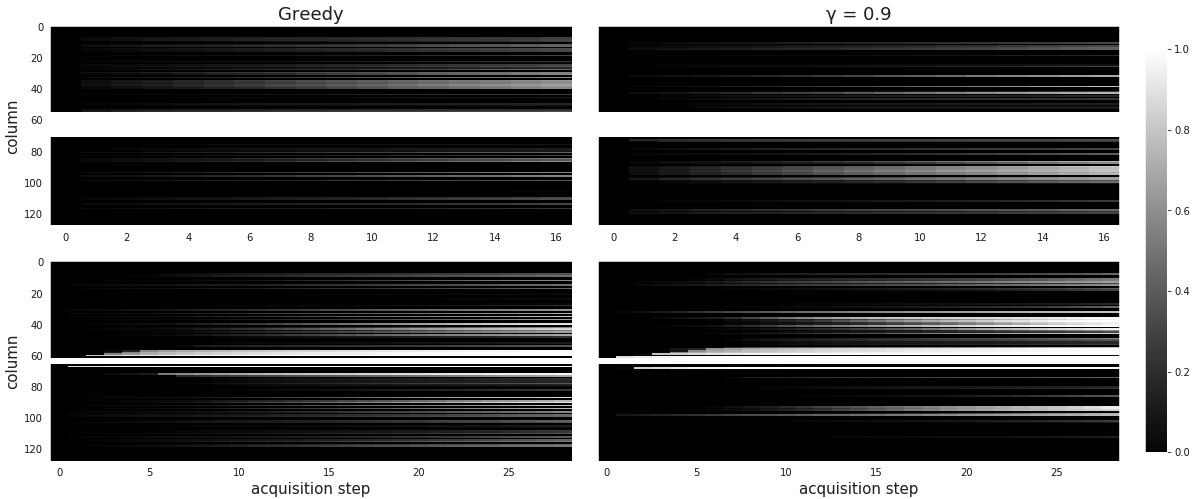}
  \caption{Visualisation of sampled trajectories for base (top) and long (bottom) horizons, averaged over the Knee test data. Shown is the fraction of MR slices for which a particular column has been sampled at an acquisition step. The central white bands are initialisation measurements. The $\gamma = 0.9$ average policies interpolate between the Greedy and NGreedy average policies. For each setting the best model on the test set was used.}
  \label{fig:polknee09}
\end{figure}

\begin{figure}
  \centering
  \includegraphics[width=.8\linewidth]{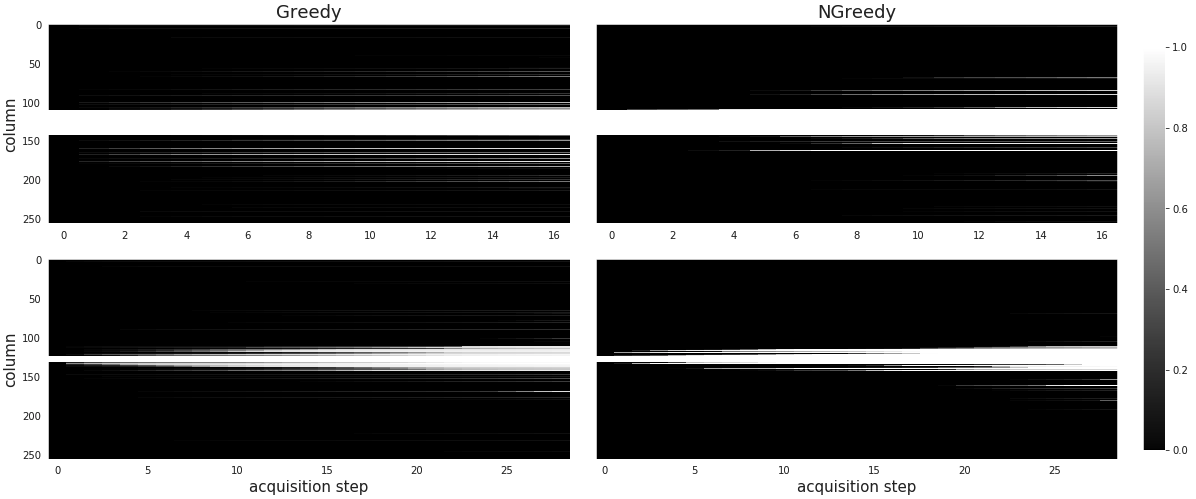}
  \caption{Visualisation of sampled trajectories for base (top) and long (bottom) horizons, averaged over the Brain test data. Shown is the fraction of MR slices for which a particular column has been sampled at an acquisition step. The central white bands are initialisation measurements. The NGreedy policies select the same column for various slices more often than the Greedy model, suggesting it adapts less to the individual images. For each setting the best model on the test set was used.}
  \label{fig:polbrain}
\end{figure}

\begin{figure}
  \centering
  \includegraphics[width=.8\linewidth]{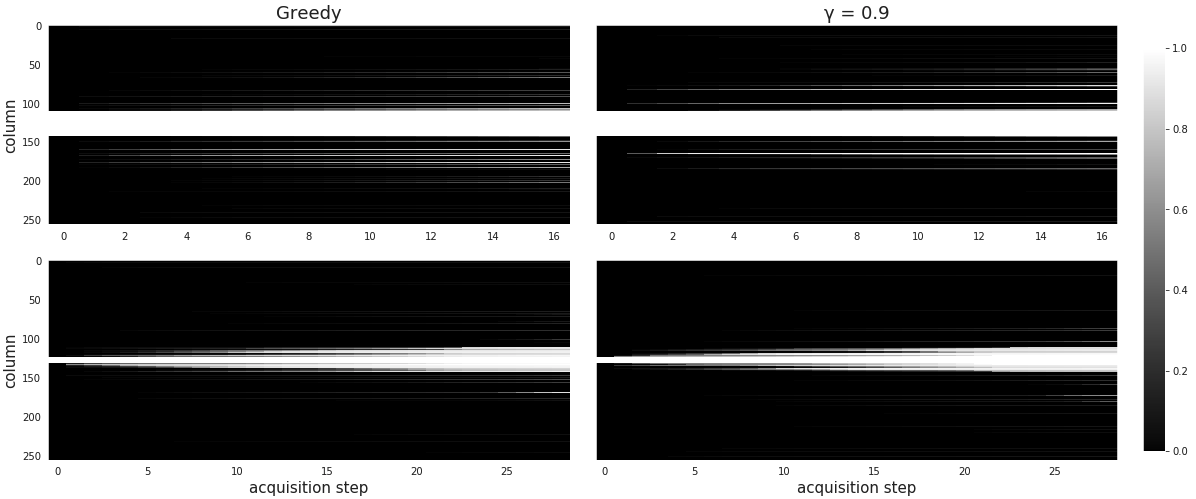}
  \caption{Visualisation of sampled trajectories for base (top) and long (bottom) horizons, averaged over the Brain test data. Shown is the fraction of MR slices for which a particular column has been sampled at an acquisition step. The central white bands are initialisation measurements. The $\gamma = 0.9$ average policies interpolate between the Greedy and NGreedy average policies. For each setting the best model on the test set was used.}
  \label{fig:polbraingamma}
\end{figure}

Here we present additional policy visualisations that were omitted from the main text. In Figure~\ref{fig:polknee09} we compare the Greedy model with the $\gamma = 0.9$ model for Knee data. The latter outputs more sharply peaked average policies than the former, although - as expected - this difference is less stark than that of Figure~\ref{fig:pol}. These visualisations were obtained by running the policy model for all MR slices in the corresponding test dataset and averaging the number of times a particular measurement was performed at a particular acquisition step. Since the acquisitions are themselves policy samples, these visualisations will slightly differ between runs of the same model.
Figure~\ref{fig:polbrain} shows a comparison of average policies of the Greedy and NGreedy model for the Brain dataset, and Figure~~\ref{fig:polbraingamma} shows the Greedy and $\gamma = 0.9$ average policies. We observe a similar contrast here as we did for Knee data, with the NGreedy model outputting more sharply peaked average policies, and the $\gamma = 0.9$ interpolating. Note that the acquisitions done by the models make up a smaller fraction of the initially selected (and total) k-space columns here than in the Knee case, due to the use of larger images.

\subsection{Learning curves}
\begin{figure}
  \centering
  \includegraphics[width=\linewidth]{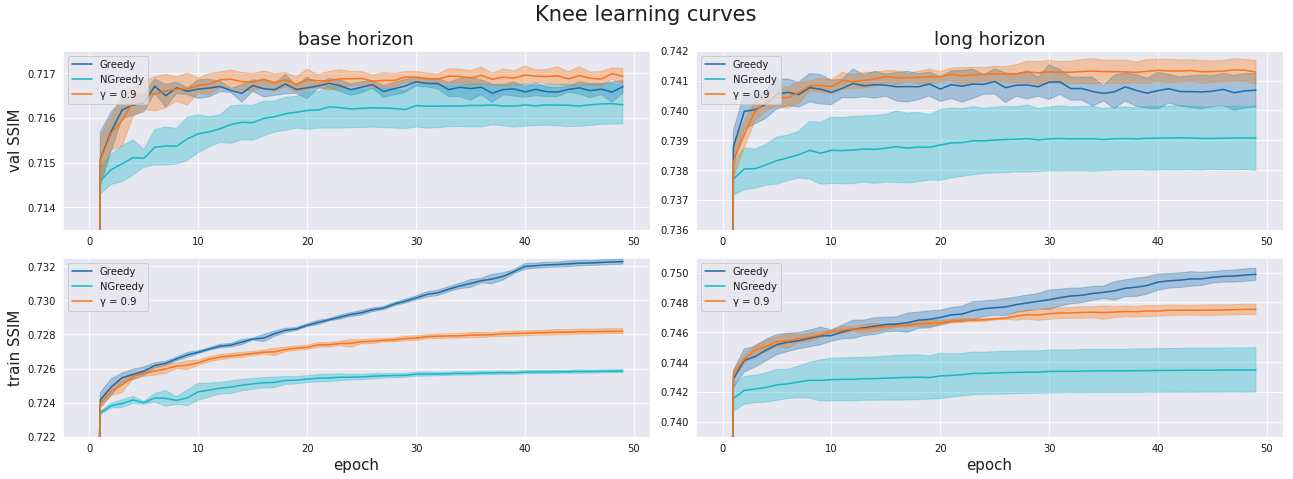}
  \caption{Learning curves on the Knee data train and validation set. For all methods the training SSIM increases steadily with the number of epochs trained, but the effect is stronger the greedier the model, suggesting the Greedy model overfits more strongly than the NGreedy model.}
  \label{fig:lckneegnggamma09}
\end{figure}

\begin{figure}
  \centering
  \includegraphics[width=\linewidth]{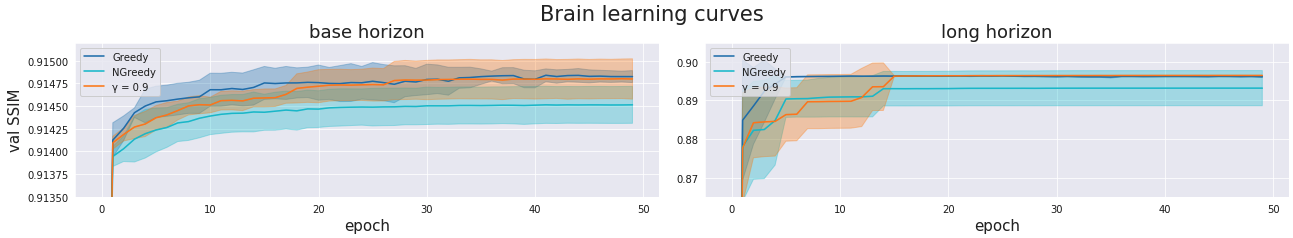}
  \caption{Learning curves on the Brain data validation set. The Greedy model converges quickest, with $\gamma = 0.9$ interpolating between Greedy and NGreedy.}
  \label{fig:lcbraingnggamma09}
\end{figure}

Figure~\ref{fig:lckneegnggamma09} provides training and validation learning curves for the Knee dataset. Note that the Greedy models seem to overfit slightly more than the NGreedy models, consistent with our hypothesis of higher gradient variance leading to the latter learning from more average reward signals.
Figure~\ref{fig:lcbraingnggamma09} provides validation learning curves for the Brain dataset. The training data learning curves were not computed for the Brain dataset, as the computation corresponds to a near doubling of the number of training epochs.

\subsection{Policy mutual information} \label{sec:appmi}
\subsubsection{Mutual information motivation} \label{sec:appmimot}
In section~\ref{sec:res} we showed the average policy over the test data set for the best Greedy and NGreedy models. This average policy $\pi(a_t)$ for acquisition step $t$ was computed by running a single trajectory for every setting and taking the average over the data set of $N$ points indexed by $i$ as: $\pi(a_t) = \frac{1}{N} \sum_{i=1}^N \pi(a_t|s_{i,t})$, where $s_{i,t}$ is the state corresponding to MR image $i$ at acquisition step $t$.

As stated, the Greedy policy seemed to be more adaptive as its average policy was less peaked that that of the NGreedy policy. However, inspection of $\pi(a_t)$ on its own is not sufficient to support this claim, as the uncertainty in the policy could also be explained as the Greedy model having high uncertainty in $\pi(a_t|s_t)$: that is, high uncertainty on which action to sample even given the current reconstruction. 

In the main text we further supported our adaptivity claim by comparing the Greedy model to a non-adaptive oracle, which could only be outperformed by being adaptive to the current state, which in turn implies low uncertainty in $\pi(a_t|s_t)$ relative to $\pi(a_t)$. Here we directly measure the gap between these two uncertainties for the Greedy and NGreedy model in both the base and long horizon settings.

We use the entropy as a quantitative measure of the uncertainty in the policies. The entropy $H(A_t)$ of a probability distribution over possible actions $A_t$ is known as the marginal entropy, computed as $H(A_t) = \sum_j \pi(a_{j,t}) \log \pi(a_{j,t})$, where $j$ indexes the actions in $A_t$. The conditional entropy $H(A_t|S_t)$ of a conditional probability distribution over actions $A_t$ given a state $s_t \in S_t$ is computed as $H(A_t|S_t) = \sum_j \pi(a_{j,t}|s_t) p(s_t) \log \pi(a_{j,t}|s_t) = \frac{1}{N} \sum_{i,j} \pi(a_{j,t}|s_{i,t}) \log \pi(a_{j,t}|s_{i,t})$, again for the $N$ data points in the test set.

The gap between these two entropies is the mutual information (MI) $I(A_t;S_t)$ of $A_t$ and $S_t$: $I(A_t;S_t) = H(A_t) - H(A_t|S_t)$. This is a quantitative measure of how much information the state gives about the action, under the learned policy. This mutual information provides a direct measure of how adaptive a model is: the higher $I(A_t;S_t)$, the more the model changes its policy as the state changes.

Note that high mutual information does not on its own equal strong performance on the MRI subsampling task. As a degenerate example, consider the case of a specific Knee policy that performs a single acquisition step on 128 slices. For every slice this policy suggests a different measurement with probability 1. The marginal entropy of this policy is $\ln(128)$, and the conditional entropy is $0$. This gives the maximum possible mutual information of $\ln(128)$ for this setting - and indeed this policy is maximally adaptive - but clearly this is a bad policy for the task at hand. Of course, as our policies are learned based on the SSIM reward signal, they are incentivised to be adaptive only if this helps the MRI task, and thus the MI gives information about the degree to which these policies are usefully adaptive.

\subsubsection{Knee data mutual information} \label{sec:kneemi}
In Figure~\ref{fig:mikneegnggamma09} of the main text we visualise the MI per acquisition step for the Greedy, NGreedy and $\gamma = 0.9$ models on Knee data. For the base horizon task, it seems the Greedy model learns to be adaptive already in the first acquisition, having enough information to produce adaptive policies. The NGreedy model has low mutual information for the initial acquisition steps, but becomes more adaptive as it gets closer to exhausting the acquisition budget. 

An explanation for the NGreedy model's behaviour may go as follows: if little is known about an MR image, one might reasonably default to taking measurements in the low-frequency bands, until enough information has been acquired to focus on image-specific details. However, as the Greedy model has high mutual information already at acquisition initialisation, it is more likely that the NGreedy model performs better at later acquisition steps due to the shorter time horizon for optimisation. This is consistent with our hypothesis that gradient variance due to long optimisation horizons hampers the NGreedy model. The NGreedy model eventually overtakes the Greedy model in mutual information: it may be that the NGreedy model has the potential to be more adaptive than the Greedy model for certain acquisition steps due to its non-greediness. Another possible explanation is that the usefulness of adaptivity decreases as more measurements have already been adaptively sampled, and as such the NGreedy model catches up to the Greedy model as the latter runs into diminishing marginal returns more quickly.

Interesting is the bowl-like shape in the Greedy model's MI. It implies higher adaptivity at the start and end of the acquisition trajectory, but whether this is due to properties of the problem setting or due to properties of training is unclear and left for future research.

\begin{figure}
  \centering
  \includegraphics[width=\linewidth]{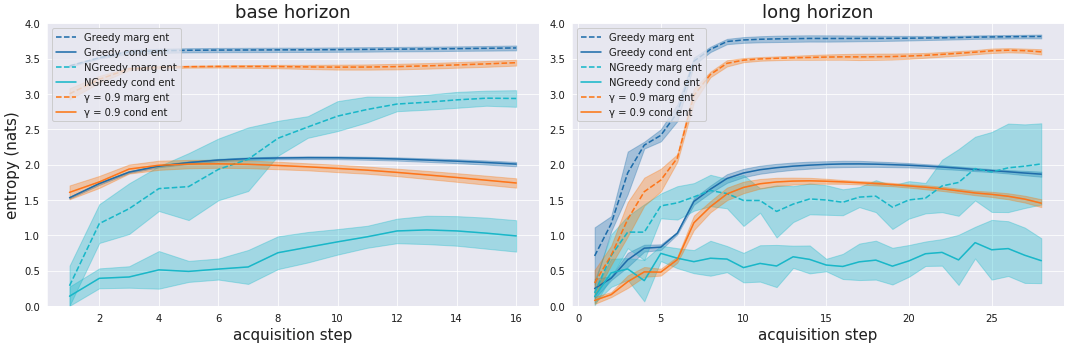}
  \caption{Marginal and conditional entropies for the base (left) and long (right) horizon settings for the Greedy, NGreedy and $\gamma = 0.9$ methods on Knee test data. Shown is the average and standard deviation of the entropies per acquisition step over five seeds, computed with $q=8$ trajectories.}
  \label{fig:entskneegnggamma09}
\end{figure}

For the long horizon task, the Greedy model's mutual information steeply climbs during the first few acquisitions, suggesting that initially it does not have enough information to properly adapt its policies to the input: this behaviour mirrors that of the NGreedy model in the base horizon case. The low mutual information means the marginal and conditional entropies are close together in value. To further analyse this behaviour, we show the conditional and mutual entropies separately in Figure~\ref{fig:entskneegnggamma09}. Since all entropies are relatively low at acquisition initialisation of the long horizon task, we conclude both models start out selecting a small set of similar measurements for most MR images initially, corresponding to more sharply peaked conditional policies. In the base horizon setting the Greedy model starts out with relatively high values for both entropies, suggesting more adaptivity for this setting.
Figure~\ref{fig:entskneegnggamma09} also shows that the NGreedy model generally seems more certain (lower conditional entropy) about its predictions than the Greedy model. This could be related to the NGreedy model learning from an average reward signal, but there are likely other explanations consistent with the current observations as well. While it might also be indicative of the NGreedy model overfitting relative to the Greedy model, we do not observe this here, as shown in Figure~\ref{fig:lckneegnggamma09}.

In the long horizon task, the Greedy model obtains the same level of adaptivity as on the base horizon task through the first few acquisitions. After this, it shows the same bowl-like MI shape observed for the base horizon task. The NGreedy model shows similar behaviour as it did on the base horizon task, presenting stronger adaptivity for the final few acquisition steps, likely due to shorter optimisation horizons. Unlike in the base horizon task, it does not catch up to the Greedy model in mutual information for any acquisition step. This provides another indication that the NGreedy model's lack of adaptivity compared to the Greedy model is correlated to longer optimisation horizons.

The behaviour of the $\gamma = 0.9$ MI and entropies is more similar to that of the Greedy model than that of NGreedy model, indicating that the learned policies exploit similar adaptivity information, while retaining a less Greedy optimisation horizon. This is somewhat surprising, as its optimisation objective and practical training details are much more similar to the NGreedy model than to the Greedy model. Interestingly, the $\gamma = 0.9$ model manages to surpass the Greedy model in mutual information near the end of the acquisition horizon for both settings, which may contribute to its higher performance.

\subsubsection{Brain data mutual information}
\begin{figure}
  \centering
  \includegraphics[width=\linewidth]{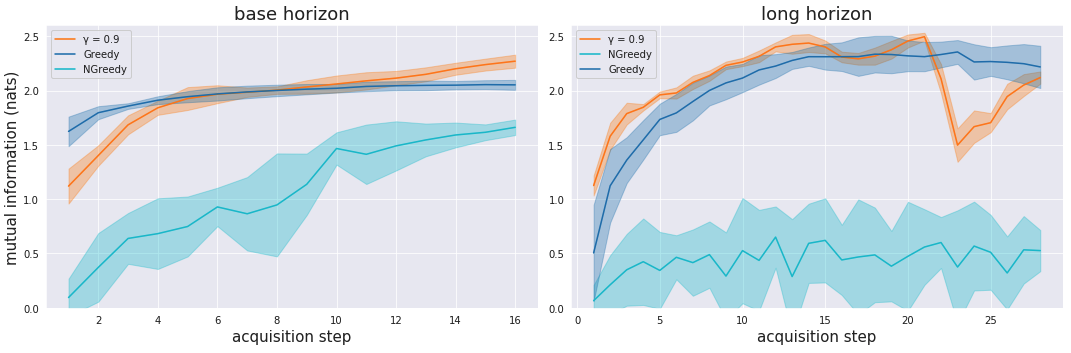}
  \caption{Mutual information for the base (left) and long (right) horizon settings, for the Greedy, NGreedy and $\gamma = 0.9$ methods on Brain data. Shown is the average and standard deviation of the mutual information per acquisition step over five seeds, computed with $q=8$ trajectories.}
  \label{fig:mibraingnggamma09}
\end{figure}

\begin{figure}
  \centering
  \includegraphics[width=\linewidth]{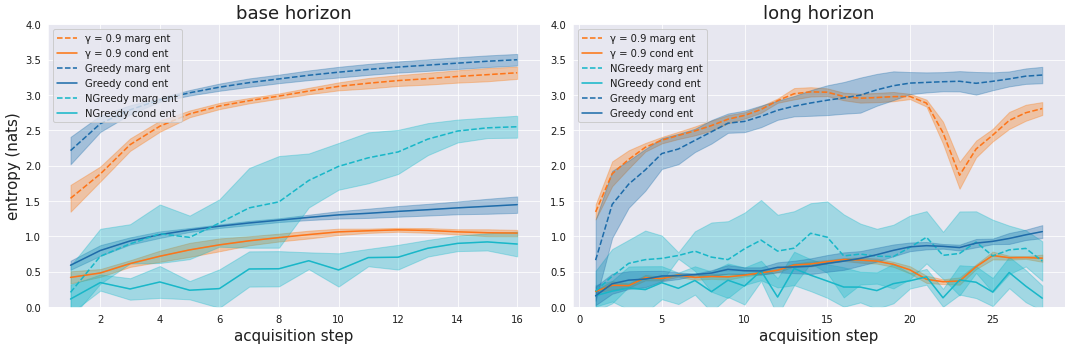}
  \caption{Mutual information for the base (left) and long (right) horizon settings, for the Greedy, NGreedy and $\gamma = 0.9$ methods on Brain data. Shown is the average and standard deviation of the mutual information per acquisition step over five seeds, computed with $q=8$ trajectories.}
  \label{fig:entbraingnggamma09}
\end{figure}

In Figures~\ref{fig:mibraingnggamma09} and~\ref{fig:entbraingnggamma09} we present the mutual information and entropy plots for the Greedy and NGreedy model trained on Brain data. The behaviour of these quantities is quite similar to their Knee data counterparts in Figures~\ref{fig:mikneegnggamma09} and~\ref{fig:entskneegnggamma09}: the Greedy model generally enjoys higher average MI than the NGreedy model, as well as lower variance. The most notable differences are the lack of bowl-like shape in the Greedy MI plots, as well as the inability of the NGreedy MI to overtake the Greedy even in the base horizon setting.
Whereas for the Knee dataset the base horizon setting corresponds to a final acceleration factor of 4, for the Brain dataset the final acceleration factor is $\frac{256}{32 + 16} = 5 \frac{1}{3}$, so this comparison is not strictly fair. We note however than for the Knee dataset base horizon setting, the NGreedy model MI overtakes that of the Greedy model already after only half the total acquisition steps have been performed, corresponding to the same acceleration factor $\frac{128}{16 + 8} = 5 \frac{1}{3}$. Indeed, looking back at Table~\ref{tab:ssims}, the relative gap in performance between the Greedy and NGreedy model is larger for the Brain dataset than for the Knee dataset.

In the long horizon setting, the MI of the Greedy model requires about twice the number of acquisition steps (compared to Knee data) to reach the point where it flattens out. Since the Brain images contains twice as many columns, this corresponds to the same relative acceleration (around $\frac{1}{12}$).

The $\gamma = 0.9$ Brain models behave broadly similarly to the Knee case, with the exception of the heavy drop and recovery of the marginal entropy late in the long horizon setting. We leave exploration of this behaviour to future work. We furthermore note that in the long horizon setting the $\gamma = 0.9$ MI starts out higher than the Greedy MI. It seems the longer optimisation horizon does not hamper the adaptivity of this model as much, perhaps due to the lower acceleration factor relative to the Knee setting.

\subsubsection{Mutual information on Knee data for various discount factors}
\begin{figure}
  \centering
  \includegraphics[width=\linewidth]{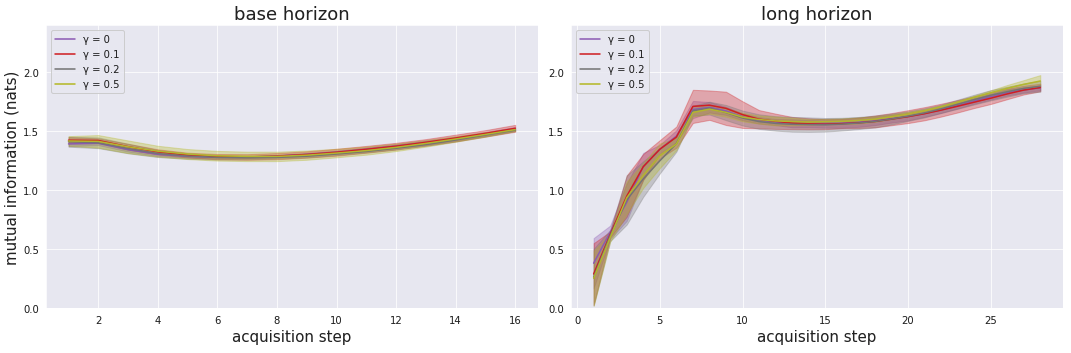}
  \caption{Mutual information for the base (left) and long (right) horizon settings for the NGreedy method with various values of the discount factor $\gamma \in [0, 0.5]$ on Knee test data. Shown is the average and standard deviation of the mutual information per acquisition step over five seeds, computed with $q=8$ trajectories.}
  \label{fig:mikneelowgamma}
\end{figure}

\begin{figure}
  \centering
  \includegraphics[width=\linewidth]{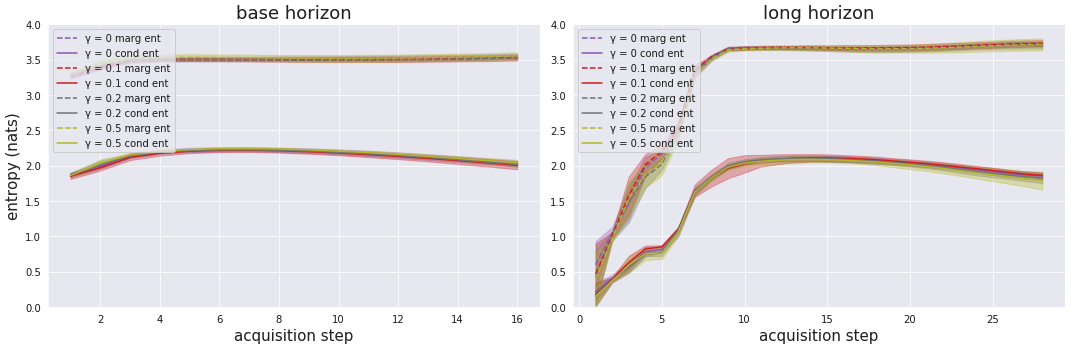}
  \caption{Marginal and conditional entropies for the base (left) and long (right) horizon settings for the NGreedy method with various values of the discount factor $\gamma \in [0, 0.5]$ on Knee test data. Shown is the average and standard deviation of the mutual information per acquisition step over five seeds, computed with $q=8$ trajectories.}
  \label{fig:entkneelowgamma}
\end{figure}

\begin{figure}
  \centering
  \includegraphics[width=\linewidth]{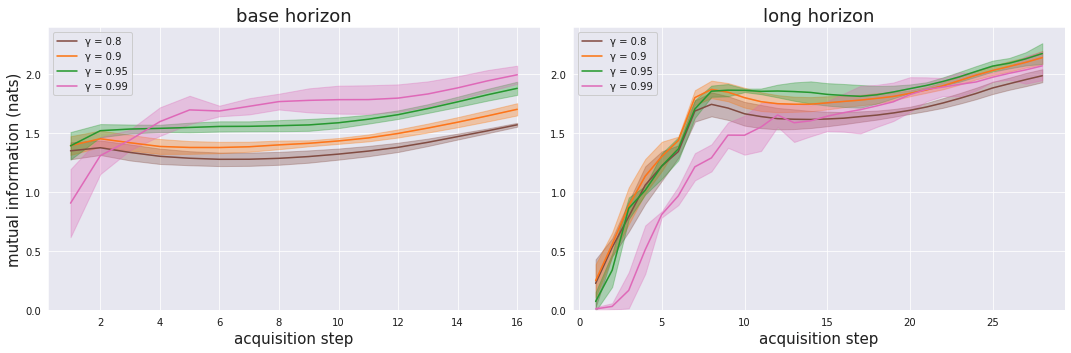}
  \caption{Mutual information for the base (left) and long (right) horizon settings for the NGreedy method with various values of the discount factor $\gamma \in [0.5, 1]$ on Knee test data. Shown is the average and standard deviation of the mutual information per acquisition step over five seeds, computed with $q=8$ trajectories.}
  \label{fig:mikneehighgamma}
\end{figure}

\begin{figure}
  \centering
  \includegraphics[width=\linewidth]{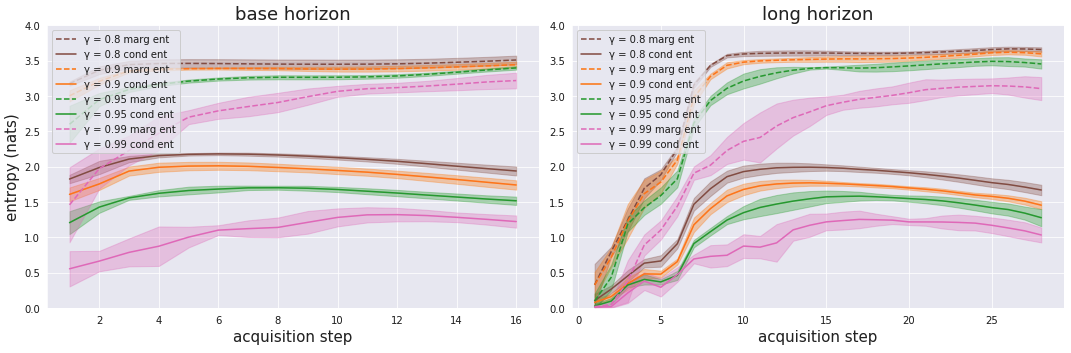}
  \caption{Marginal and conditional entropies for the base (left) and long (right) horizon settings for the NGreedy method with various values of the discount factor $\gamma \in [0.5, 1]$ on Knee test data. Shown is the average and standard deviation of the mutual information per acquisition step over five seeds, computed with $q=8$ trajectories.}
  \label{fig:entkneehighgamma}
\end{figure}

In Figures~\ref{fig:mikneelowgamma},~\ref{fig:mikneehighgamma},~\ref{fig:entkneelowgamma}, and~\ref{fig:entkneehighgamma}, we present mutual information, conditional entropy, and marginal entropy plots for NGreedy models trained on Knee data with various values of the discount factor $\gamma$. These models represent an interpolation between the Greedy and NGreedy model. See section~\ref{sec:appssim} for SSIM performance of these models. Note that the $\gamma = 0$ model does not correspond to the Greedy model due to a difference in computation of the reward baseline (see Figure~\ref{fig:psamp}).

In Table~\ref{tab:ssimsgamma}, NGreedy models with $\gamma \in [0, 0.5]$ all perform quite similarly to the Greedy model, and indeed their MI and entropy curves look very similar to the Greedy model's. For $\gamma \in [0.5, 1]$ the MI and entropy curves look more like interpolations between the Greedy and NGreedy case for both horizon settings. However, the behaviour surprisingly still seems more similar to the Greedy model than the NGreedy model even for $\gamma = 0.99$.

\subsection{Additional SSIM results} \label{sec:appssimsup}
\subsubsection{Extended SSIM table}
We report extended SSIM results in Table~\ref{tab:ssimsoracle}. Here we include results for a greedy oracle model, that selects the measurement leading to the greatest immediate SSIM improvement for every slice separately. We denote this model as `Oracle'. In principle this is an upper bound on any greedy model, but note that this method is most susceptible to failing to identify situations where a combination of two measurements that are separately uninformative lead to strong improvements. This seems to be the likely explanation for the low Oracle performance in the long horizon Brain setting. Nevertheless, these scores provide an indication of a performance gap that may still be closed by future research.

Additionally, we include a comparison with equispaced masks, which are generally easier to implement in MRI machines than random masks, and may perform better as noted in \citep{ete}. As with the random baseline, we initialise these masks by sampling low frequency bands up to the starting acceleration. The two-sided equispaced mask `Equi (two)' is then constructed by sampling every $r$'th column of the remaining k-space, where $r$ is determined by dividing the number of initially unsampled columns by the number of acquisitions to be made. The one-sided equispaced mask `Equi (one)' is constructed similarly, but only considering the remaining columns on one side of k-space. One-sided sampling can be more efficient due to k-space symmetry in some cases, as we observe in Table~\ref{tab:ssims}.
\begin{table}
    \caption{SSIM performance on test data. For non-deterministic models, averages and standard deviations are computed over five seeds, using $q=8$ trajectories for policy models (AlphaZero scores are averaged over three seeds instead).}
    \label{tab:ssimsoracle}
    \centering
    \begin{tabular}{lcccc}
        \toprule
         & \multicolumn{2}{c}{\textbf{Knee}} & \multicolumn{2}{c}{\textbf{Brain}} \\
         \cmidrule(r){2-3} \cmidrule(r){4-5}
         & \textbf{Base horizon} & \textbf{Long horizon} & \textbf{Base horizon} & \textbf{Long horizon} \\
         \midrule
         \textbf{Random} & $0.6948\!\pm\!0.0003$ & $0.6602\!\pm\!0.0006$ & $0.9020\!\pm\!0.0001$ & $0.5820\!\pm\!0.0006$ \\
         \textbf{Equi (one)} & $0.7049$ & $0.6880$ & $0.9038$ & $0.5862$ \\
         \textbf{Equi (two)} & $0.7064$ & $0.6918$ & $0.9016$ & $0.6026$ \\
         \textbf{NA Oracle} & $0.7213$ & $0.7421$ & $0.9099$ & $0.8909$ \\
         \textbf{Oracle} & $0.7379$ & $0.7623$ & $0.9141$ & $0.8872$ \\
         \textbf{AlphaZero} & $0.7203 \pm 0.0008$ & $0.7403 \pm 0.0009$ & - & - \\
         \textbf{NGreedy} & $0.7223 \pm 0.0003$ & $0.7421 \pm 0.0014$ & $0.9103 \pm 0.0002$ & $0.8886 \pm 0.0048$ \\
         \textbf{Greedy} & $0.7230 \pm 0.0001$ & $0.7442 \pm 0.0007$ & $0.9106 \pm 0.0001$ & $0.8917 \pm 0.0002$ \\
         $\bm{\gamma = 0.9}$ & $0.7232 \pm 0.0002$ & $0.7449 \pm 0.0004$ & $0.9106 \pm 0.0003$ & $0.8921 \pm 0.0001$ \\
        \bottomrule
    \end{tabular}
\end{table}

\subsubsection{Knee data SSIM results for other discount factors} \label{sec:appssim}
In Table~\ref{tab:ssimsgamma} we report average SSIM performance for NGreedy models trained on Knee data with various discount factors $\gamma \in [0, 1]$. These models represent an interpolation between the Greedy and NGreedy model. As noted in the main text, the model with $\gamma = 0.9$ performs best. Note the relatively heavy drop in performance very close to $\gamma = 1.0$, which corresponds to our NGreedy model. Out of all the tested discount factors it seems that a fully NGreedy ($\gamma = 1.0$) model performs worst on the Knee dataset.
\begin{table}
    \caption{Average SSIM performance on Knee test data for NGreedy models trained with various $\gamma \in [0, 1]$, the NGreedy model reported in the main text ($\gamma = 1.0$) and the Greedy model. Averages and standard deviations are computed over five seeds, with $q =8$ trajectories.}
    \label{tab:ssimsgamma}
    \centering
    \begin{tabular}{lcc}
        \toprule
         & \multicolumn{2}{c}{\textbf{Knee}} \\
         \cmidrule(r){2-3}
         & \textbf{Base horizon} & \textbf{Long horizon} \\
         \midrule
         \textbf{Greedy} & $0.7230 \pm 0.0001$ & $0.7442 \pm 0.0007$ \\
         $\bm{\gamma = 0.0}$ & $0.7231 \pm 0.0002$ & $0.7447 \pm 0.0002$ \\
         $\bm{\gamma = 0.1}$ & $0.7230 \pm 0.0001$ & $0.7446 \pm 0.0010$ \\
         $\bm{\gamma = 0.2}$ & $0.7230 \pm 0.0001$ & $0.7445 \pm 0.0003$ \\
         $\bm{\gamma = 0.5}$ & $0.7230 \pm 0.0002$ & $0.7443 \pm 0.0006$ \\
         $\bm{\gamma = 0.8}$ & $0.7231 \pm 0.0002$ & $0.7445 \pm 0.0004$ \\
         $\bm{\gamma = 0.9}$ & $0.7232 \pm 0.0002$ & $0.7449 \pm 0.0004$ \\
         $\bm{\gamma = 0.95}$ & $0.7232 \pm 0.0001$ & $0.7446 \pm 0.0002$ \\
         $\bm{\gamma = 0.99}$ & $0.7228 \pm 0.0004$ & $0.7437 \pm 0.0007$ \\
         \textbf{NGreedy} & $0.7223 \pm 0.0003$ & $0.7421 \pm 0.0014$ \\
        \bottomrule
    \end{tabular}
\end{table}

Note that the $\gamma = 0$ model does not exactly correspond to the Greedy model due to a difference in computation of the reward baseline (see Figure~\ref{fig:psamp}), but is otherwise equivalent. We note here that the performance of this model is slightly higher than that of the Greedy model, though still within one standard deviation. The weaker baseline of the $\gamma = 0$ model, may in fact help the optimisation by escaping local minima, as these Greedy models tend to have high SNR and thus seem to be less hampered by variance than the models with higher values for $\gamma$. The different reward baseline may matter more for the non-greedy models, as they seem more troubled by gradient variance. Valuable future research may be to investigate methods for adapting this baseline to the non-greedy case, for instance by incorporating value function learning to estimate the return of a particular node in Figure~\ref{fig:psamp}. Due to computational constraints, we do not report a table like Table~\ref{tab:ssimsgamma} for Brain data.

While the Greedy model underperforms most of the non-greedy models, the differences are in most cases within one standard deviation of performance. As the Greedy model is much less computationally intensive, the greedy approach may be favoured for certain MRI tasks. 

\subsection{More SNR results} \label{sec:appsnr}
We report SNR values as in Table~\ref{tab:snr} for additional intermediate training stages in Table~\ref{tab:snrfull}. These results are consistent with the conclusions stated in the main text. Shortly after initialisation all Knee settings have relatively similar SNR, likely due to the average reward signal dominating all settings under the initial random policy. For the Greedy estimators, it is notable that the SNR seems to rise and fall quite sharply. We suspect this effect is related to convergence, and we leave closer investigation of it to future work. 
\begin{table}
    \caption{Signal-to-Noise ratio comparison of the Greedy, NGreedy and $\gamma = 0.9$ models on both datasets, for the two time horizons trained on. Displayed are average SNR estimates obtained for the best performing model (on test data) on three runs over the train data for every setting, at various points during its training: Epoch $n$ refers to the model after the $n$'th epoch of training is completed. SNR is estimated using gradients for the final layer of the policy network, with $q=16$ samples and batch size $16$ in all settings.}
    \label{tab:snrfull}
    \centering
    \begin{tabular}{lcccccc}
        \toprule
         & \multicolumn{6}{c}{\textbf{Knee}} \\
         & \multicolumn{3}{c}{\textbf{Base horizon}} & \multicolumn{3}{c}{\textbf{Long horizon}} \\
         \cmidrule(r){2-4} \cmidrule(r){5-7}
         & \textbf{Greedy} & \textbf{NGreedy} & $\bm{\gamma = 0.9}$ & \textbf{Greedy} & \textbf{NGreedy} & $\bm{\gamma = 0.9}$ \\
         \midrule
        \textbf{Epoch 1} & $2.21\!\pm\!0.24$ & $1.82\!\pm\!0.01$ & $2.22\!\pm\!0.18$ & $2.46\!\pm\!0.25$ & $1.68\!\pm\!0.14$ & $2.05\!\pm\!0.11$ \\
        \textbf{Epoch 10} & $3.20\!\pm\!0.10$ & $1.17\!\pm\!0.03$ & $2.00\!\pm\!0.22$ & $5.90\!\pm\!0.14$ & $1.16\!\pm\!0.18$ & $2.00\!\pm\!0.05$ \\
        \textbf{Epoch 20} & $3.91\!\pm\!0.08$ & $1.24\!\pm\!0.07$ & $2.12\!\pm\!0.06$ & $3.49\!\pm\!0.27$ & $1.04\!\pm\!0.16$ & $3.04\!\pm\!0.06$ \\
        \textbf{Epoch 30} & $5.90\!\pm\!0.12$ & $1.08\!\pm\!0.20$ & $2.20\!\pm\!0.12$ & $4.78\!\pm\!0.13$ & $1.04\!\pm\!0.20$ & $2.24\!\pm\!0.21$ \\
        \textbf{Epoch 40} & $2.80\!\pm\!0.40$ & $1.02\!\pm\!0.12$ & $1.59\!\pm\!0.03$ & $2.14\!\pm\!0.08$ & $1.06\!\pm\!0.08$ & $1.51\!\pm\!0.12$ \\
        \textbf{Epoch 50} & $2.51\!\pm\!0.03$ & $1.02\!\pm\!0.07$ & $1.43\!\pm\!0.10$ & $2.15\!\pm\!0.16$ & $0.96\!\pm\!0.16$ & $1.29\!\pm\!0.11$ \\
        \midrule
         & \multicolumn{6}{c}{\textbf{Brain}} \\
         & \multicolumn{3}{c}{\textbf{Base horizon}} & \multicolumn{3}{c}{\textbf{Long horizon}} \\
         \cmidrule(r){2-4} \cmidrule(r){5-7}
         & \textbf{Greedy} & \textbf{NGreedy} & $\bm{\gamma = 0.9}$ & \textbf{Greedy} & \textbf{NGreedy} & $\bm{\gamma = 0.9}$ \\
         \midrule
        \textbf{Epoch 1} & $6.70\!\pm\!0.09$ & $3.76\!\pm\!0.22$ & $5.31\!\pm\!0.10$ & $8.75\!\pm\!0.20$ & $1.57\!\pm\!0.11$ & $7.80\!\pm\!0.10$ \\
        \textbf{Epoch 10} & $8.44\!\pm\!0.06$ & $2.48\!\pm\!0.11$ & $4.91\!\pm\!0.08$ & $12.10\!\pm\!0.14$ & $1.96\!\pm\!0.07$ & $6.28\!\pm\!0.17$ \\
        \textbf{Epoch 20} & $11.21\!\pm\!0.08$ & $2.95\!\pm\!0.23$ & $5.32\!\pm\!0.02$ & $13.36\!\pm\!0.19$ & $1.18\!\pm\!0.15$ & $7.22\!\pm\!0.19$ \\
        \textbf{Epoch 30} & $14.50\!\pm\!0.09$ & $2.02\!\pm\!0.21$ & $3.25\!\pm\!0.05$ & $16.16\!\pm\!0.18$ & $1.15\!\pm\!0.12$ & $6.50\!\pm\!0.07$  \\
        \textbf{Epoch 40} & $14.91\!\pm\!0.11$ & $1.83\!\pm\!0.14$ & $3.08\!\pm\!0.23$ & $13.04\!\pm\!0.04$ & $1.04\!\pm\!0.14$ & $3.73\!\pm\!0.26$ \\
        \textbf{Epoch 50} & $7.02\!\pm\!0.07$ & $1.45\!\pm\!0.10$ & $2.35\!\pm\!0.00$ & $4.56\!\pm\!0.09$ & $0.82\!\pm\!0.08$ & $2.98\!\pm\!0.09$ \\
        \bottomrule
    \end{tabular}
\end{table}

\subsection{Reconstruction examples}
\begin{figure}
  \centering
  \includegraphics[width=\linewidth]{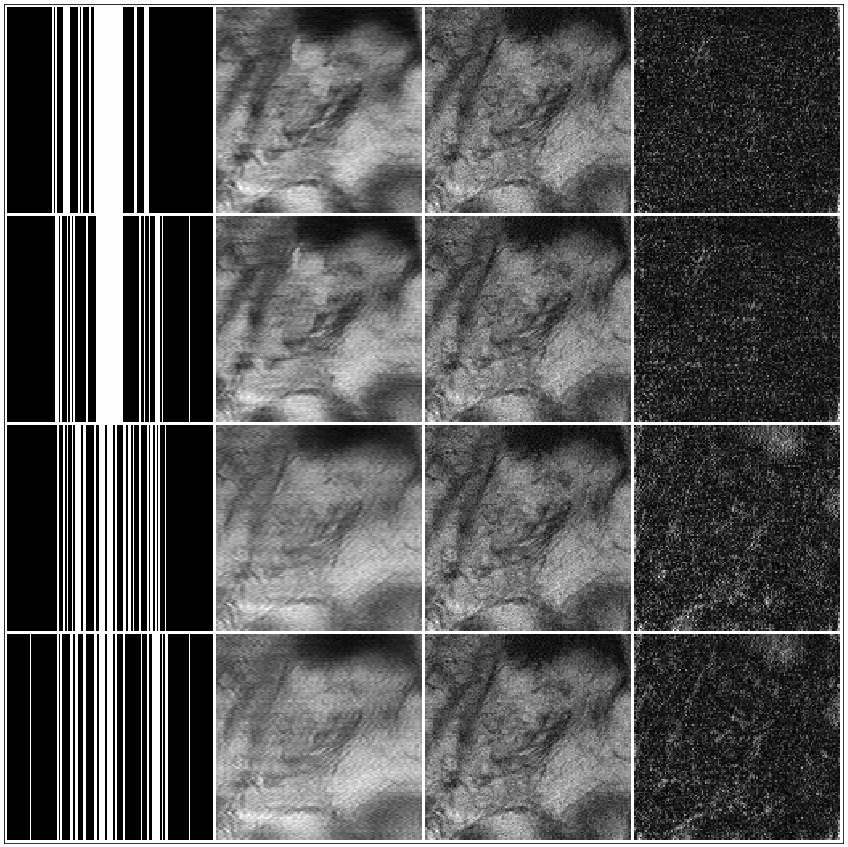}
  \caption{Visualisation of a slice of Knee data for various settings. From top to bottom: base horizon Greedy, base horizon NGreedy, long horizon Greedy, long horizon NGreedy. From left to right: final subsampling mask, reconstruction at this point, target image, absolute difference between target and reconstruction.}
  \label{fig:visknee1}
\end{figure}

\begin{figure}
  \centering
  \includegraphics[width=\linewidth]{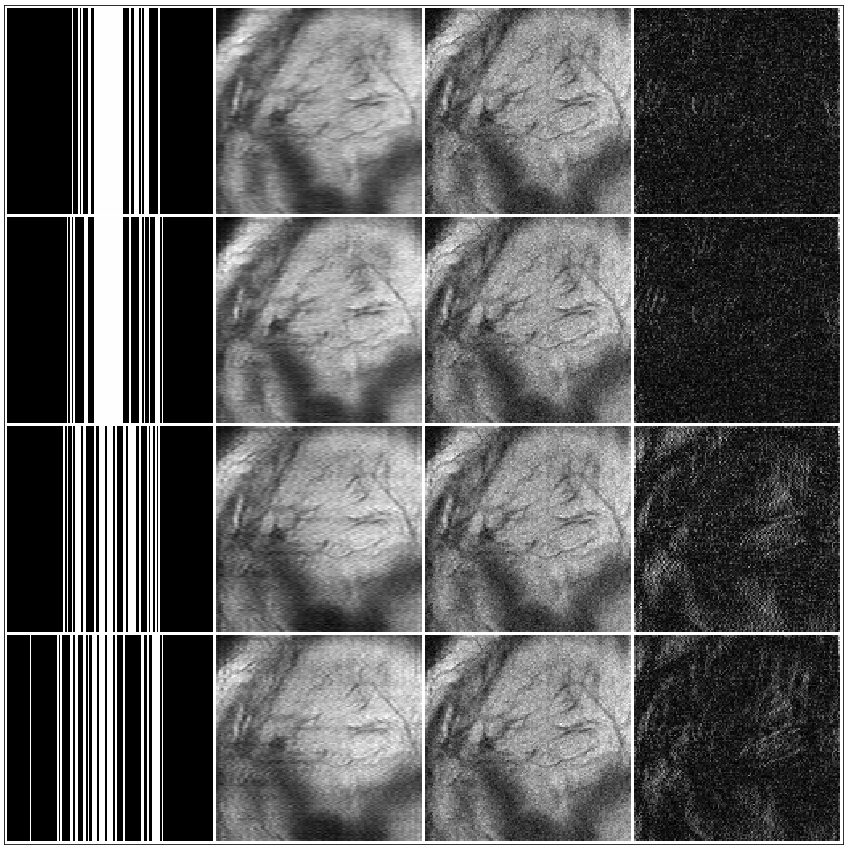}
  \caption{Visualisation of a slice of Knee data for various settings. From top to bottom: base horizon Greedy, base horizon NGreedy, long horizon Greedy, long horizon NGreedy. From left to right: final subsampling mask, reconstruction at this point, target image, absolute difference between target and reconstruction.}
  \label{fig:visknee70}
\end{figure}

\begin{figure}
  \centering
  \includegraphics[width=\linewidth]{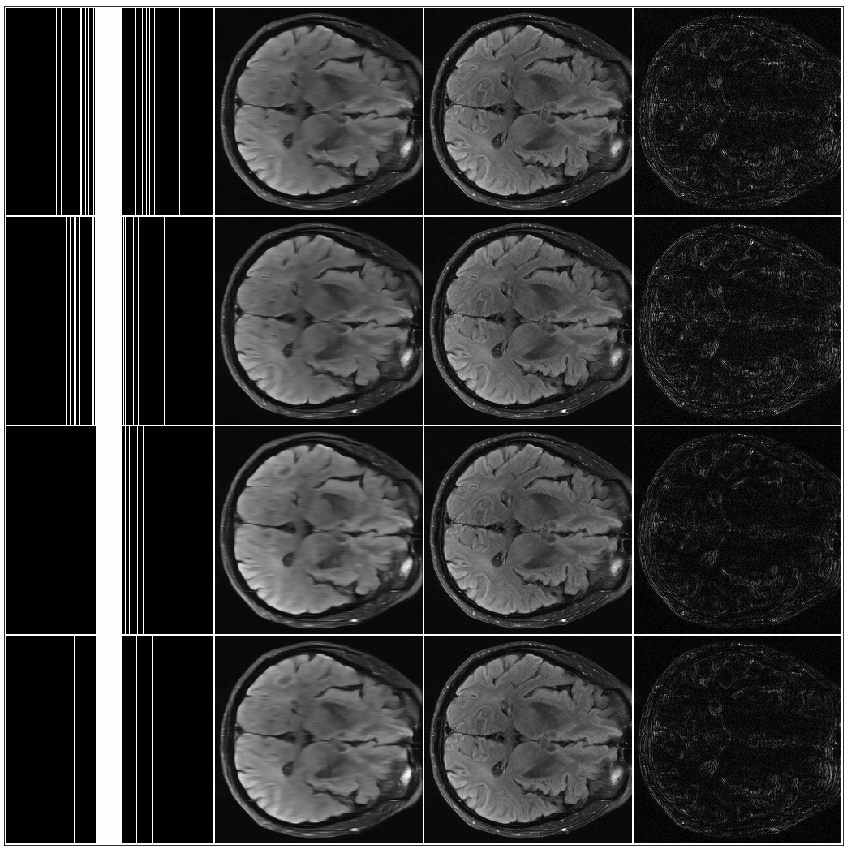}
  \caption{Visualisation of a slice of Brain data for various settings. From top to bottom: base horizon Greedy, base horizon NGreedy, long horizon Greedy, long horizon NGreedy. From left to right: final subsampling mask, reconstruction at this point, target image, absolute difference between target and reconstruction.}
  \label{fig:visbrain1}
\end{figure}

\begin{figure}
  \centering
  \includegraphics[width=\linewidth]{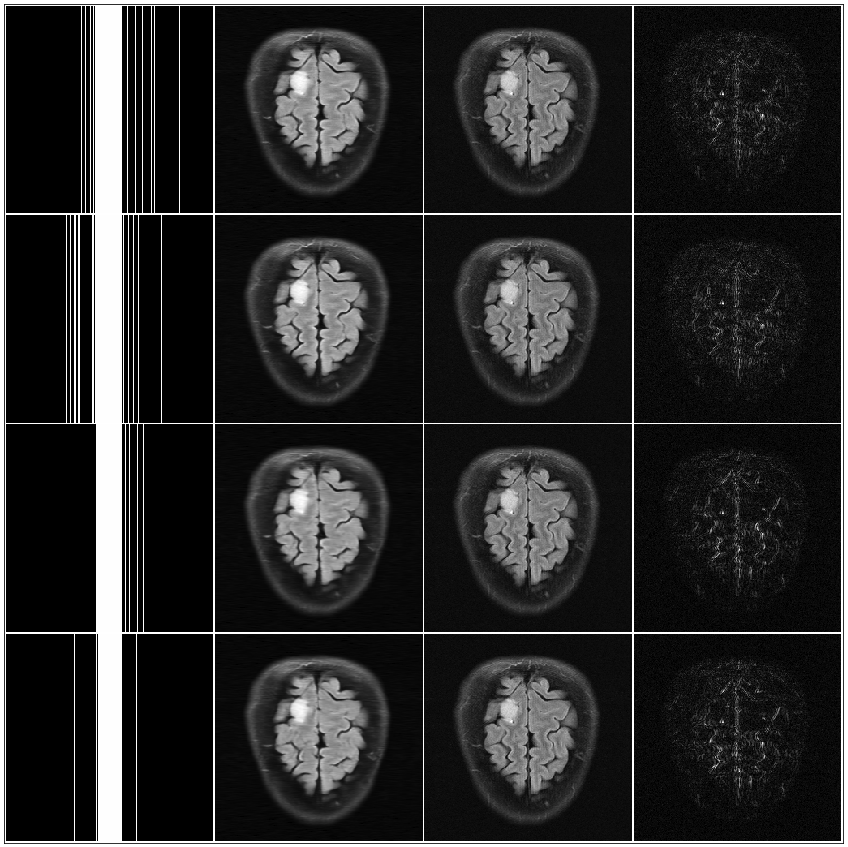}
  \caption{Visualisation of a slice of Brain data for various settings. From top to bottom: base horizon Greedy, base horizon NGreedy, long horizon Greedy, long horizon NGreedy. From left to right: final subsampling mask, reconstruction at this point, target image, absolute difference between target and reconstruction.}
  \label{fig:visbrain42}
\end{figure}

Here we present some example reconstructions for the Greedy and NGreedy model on both acquisition horizons. Figures~\ref{fig:visknee1} and~\ref{fig:visknee70} each show a slice of knee data. Figures~\ref{fig:visbrain1} and~\ref{fig:visbrain42} instead each show a slice of brain data. 

In every image we present from top to bottom: base horizon Greedy, base horizon NGreedy, long horizon Greedy, long horizon NGreedy. From left to right: final subsampling mask, reconstruction at this point, target image, absolute difference between target and reconstruction. For the Brain data long horizon setting, the policies primarily suggest to sample low frequency bands, as was already observed in Figure~\ref{fig:polbrain}.


\end{document}